\documentclass[journal]{IEEEtran}

\usepackage{xcolor}
\usepackage{graphicx}
\usepackage{comment}
\usepackage{amssymb}
\usepackage{amsmath}
\usepackage{caption}
\usepackage{array,multirow}

\newcommand{\Matthew}[1]{}
\newcommand{\Tiago}[2]{}
\newcommand{\Adam}[3]{}
\newcommand{\Anurag}[4]{}
\begin{document}
%
\title{Procedural Content Generation via Knowledge Transformation (PCG-KT)}
%
%
%
\definecolor{matthewcolor}{rgb}{0.0,0.2,0.8}
\newcommand{\Guzdial}[1]{{\color{matthewcolor}{MG: #1}}}


\author{Anurag Sarkar, Matthew Guzdial, Sam Snodgrass, Adam Summerville, Tiago Machado, Gillian Smith
\thanks{A. Sarkar is with the Khoury College of Computer Sciences, Northeastern University, Boston, MA, USA (email: sarkar.an@northeastern.edu)}
\thanks{M. Guzdial is with the Department of Computing Science, Alberta Machine Learning Institute, University of Alberta, Edmonton, AB, Canada (email: guzdial@ualberta.ca)}
\thanks{S. Snodgrass is with modl.ai, Copenhagen, Denmark (email: sam@modl.ai)}
\thanks{A. Summerville is with the College of Science, California State Polytechnic University, Pomona, CA, USA (email: asummerville@cpp.edu)}
\thanks{T. Machado is with Northeastern University, Boston, MA, USA (email:tiago.la.machado@gmail.com)}
\thanks{G. Smith is with the Department of Computer Science, Worcester Polytechnic Institute, Worcester, MA, USA (email:gmsmith@wpi.edu)}
}

%
%

\markboth{IEEE Transactions on Games}%
{Shell \MakeLowercase{\textit{et al.}}: Bare Demo of IEEEtran.cls for IEEE Journals}
%



\maketitle

\begin{abstract}
We introduce the concept of Procedural Content Generation via Knowledge Transformation (PCG-KT), a new lens and framework for characterizing PCG methods and approaches in which content generation is enabled by the process of knowledge transformation---transforming knowledge derived from one domain in order to apply it in another. 
Our work is motivated by a substantial number of recent PCG works that focus on generating novel content via repurposing derived knowledge. 
Such works have involved, for example, performing transfer learning on models trained on one game's content to adapt to another game's content, as well as recombining different generative distributions to blend the content of two or more games.
Such approaches arose in part due to limitations in PCG via Machine Learning (PCGML) such as producing generative models for games lacking training data and generating content for entirely new games. 
In this paper, we categorize such approaches under this new lens of PCG-KT by offering a definition and framework for describing such methods and surveying existing works using this framework. 
Finally, we conclude by highlighting open problems and directions for future research in this area.
\end{abstract}


%

\section{Introduction}

Procedural content generation (PCG) describes the set of approaches for generating game content via algorithmic means.
There are several lenses through which we describe and understand PCG including search-based PCG (SBPCG)~\cite{togelius2010search}, PCG as a constructive method \cite{risi2020increasing}, PCG as a representation of values \cite{smith2017we}, and PCG as a means for providing different player experiences~\cite{yannakakis2011experience}.
Each of these lenses has been informed primarily by a focus on formalizing and generating content for a single game.

A recent interest in the application of machine learning to procedural content generation (PCGML~\cite{summerville2018procedural}) has revealed not only a new technical approach to content generation but also new research questions and trajectories informed by a data-oriented approach to content generation. 
PCGML approaches generally focus on learning a distribution over existing content and then sampling from that distribution to produce new content.
However, PCGML models can only generate content from within a distribution learned from their training data.
This has led to certain core problems in PCGML namely, how do we create new game content for a game with little or no training data and how can we produce entirely new content that references existing content but that is distinct from it?
Such problems give rise to further questions and avenues of work. For example, generating new content that inherits important characteristics from existing content without directly repeating or plagiarizing it, or recombining knowledge derived from existing content to generate new content and discover new game and level designs.

We identify a number of potential, longstanding challenges in PCG that can be addressed by answering the above questions. 
One is automated game design i.e., the production of entirely novel video games. 
While there have been prior approaches to this problem \cite{cook2016angelina}, they are still relatively few in number. 
This is in part due to the large amount of design knowledge required to produce such a system \cite{guzdial2018automated}. 
If we could transform knowledge from existing games towards the production of new games, we could sidestep this issue. 

Controllability stands as another core PCG problem, with the focus on how users might best achieve their desired goals with a PCG system. 
A common problem when it comes to controllability is that a procedural content generator is unable to meet a user's needs \cite{guzdial2019friend}.
For example, a user may wish to generate some content $X$. 
If the closest the PCG system can get is to generate some content $Y$, then the user will be unable to satisfactorily control the system.
But if the system had the ability to transform $Y \rightarrow X$, then this problem could be resolved.

\begin{figure*}[tb]
  \centering
  \includegraphics[width=0.9\linewidth]{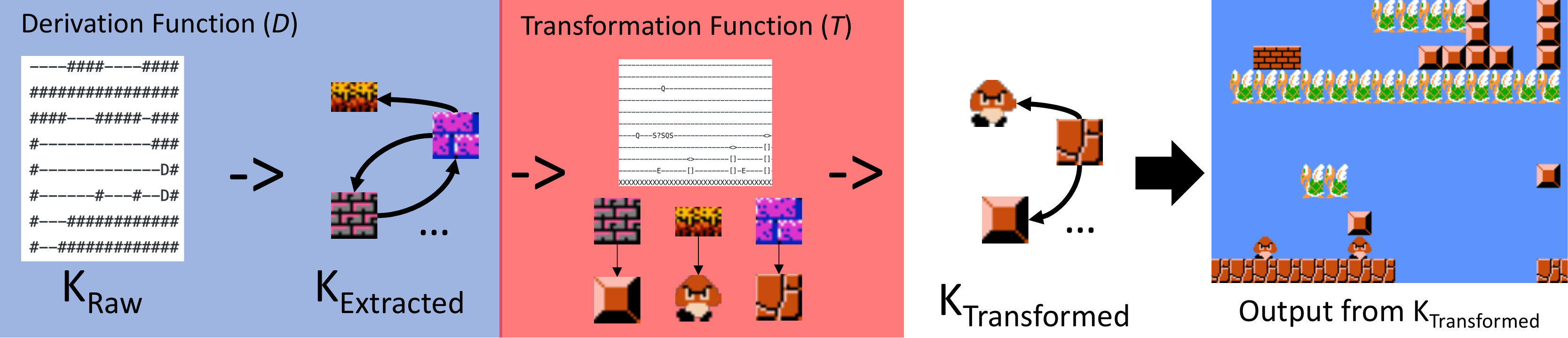}
  \caption{Example of our framework applied to an example from \cite{snodgrass2016approach}. Initial \textit{Kid Icarus} level data (raw knowledge $\mathbb{G}_1$) is used to derive a Markov level design model (extracted knowledge $\mathbb{K}_1$) via a derivation function ($D$). This knowledge is passed as input through a transformation function ($T$), which uses a mapping and a Mario level to derive a Markov level design model for \textit{Super Mario Bros.} levels (transformed knowledge $\mathbb{K}_2$), which can be used to output novel content.}
  \label{fig:ktpcgExample}
\end{figure*}

In this article, with a view towards addressing such problems, we describe a new lens for characterizing PCG research and practice: Procedural Content Generation via Knowledge Transformation (PCG-KT).
We provide a framework and vocabulary for describing PCG systems by the methods and extent to which they transform knowledge between domains.
We further survey existing systems that are situated in novel positions within this framework, and pose new potential areas of PCG research that are uncovered by this framing.

We would like to note that we introduce PCG-KT as a lens for understanding PCG processes, not as a category of PCG approaches. 
Unlike SBPCG or PCGML, we are not identifying a group of PCG approaches related by an underlying AI technology. 
Instead, we use this lens and framework to allow us to compare and contrast PCG approaches based on the extent to which they involve the transformation of knowledge. 
All PCG approaches could be understood as falling within PCG-KT, but most involve zero or trivial transformation.
On the other end of the spectrum, there are approaches that radically transform the derived knowledge to approximate secondary or nonexistent knowledge. 
For clarity in introducing these concepts, we focus on the latter of these in this paper. 
However, we view PCG-KT as a descriptive framework, not as a prescriptive set of conditions determining sufficient knowledge transformation to qualify as PCG-KT.

\section{PCG via Knowledge Transformation}

In this section, we outline nomenclature and a generalized framework for PCG via Knowledge Transformation (PCG-KT). 
We acknowledge that PCG is broad and that there are many different kinds of content, knowledge, and transformation processes used in the field. 
We refer the reader to Section \ref{sec:framework} for a framework for differentiating between types of knowledge transformation, as well as examples of how existing systems fit into this framework. 

We provide a formalized definition for PCG-KT in order to identify prior work, to prompt reflection on how this lens both fits and reframes existing approaches to PCG, and to encourage future work in under-explored directions. 

For our definition we employ the following vocabulary:
\begin{itemize}
\item \textbf{Games} ($\mathbb{G}$: a set of games; $G$: a game $G \subseteq \mathbb{G}$). A game, or set of games, forms the domain that situates and contextualizes the knowledge transformation process. We deliberately do not adopt any one prescriptive definition of ``game’’ here: a game might be theoretical, in-development, or published; renowned or unknown; a research ``toy’’ or established; digital or analog; procedurally generated or fully authored by a human. The nature of the game(s) informs the types of knowledge that can be derived, and the broader context and cultural situation of that knowledge.
\item \textbf{Knowledge} ($\mathbb{K}$: a set of knowledge derived from a set of games; $K$: a unit of knowledge). We use the term ``knowledge’’ to describe information derived from a set of games that can be applied to a new set of games (with or without transformation). This information could take many different forms, including authored content (sprites, level maps), functional design aspects (e.g., mechanics, game states), or broader information related to overall game experience (e.g., inferred style heuristics, automated playtraces). The knowledge may vary in scope (e.g., a single sprite or an entire ruleset). It may also be represented in a form that is independent of what either the original or target games use to function. For example, the ASCII representation used in level maps in the Video Game Level Corpus (VGLC) \cite{VGLC} is knowledge that is derived from those games, yet not in the format those games use (and indeed, elides some features of the original level maps such as non-interactable decorations).
\item \textbf{Derivation} (A derivation function $D$ that produces knowledge from a game: $D(\mathbb{G}_i) = \mathbb{K}_i$). Knowledge should be derived from some set of games according to a reproducible process. This knowledge may be derived by hand, automatically, or via some hybrid process. The derivation process does not need to apply directly to the game as an artifact (e.g., spritesheets extracted from game ROMs). For example, knowledge derived through inference from playtraces or gameplay video still follows a clear derivation process related to the game. 
\item \textbf{Transformation} (A transformation function $T$ that takes knowledge as input and produces knowledge as output: $T(\mathbb{K}_i) = \mathbb{K}_j$). 
This function accepts knowledge that has been derived from a set of games and alters it in such a way that it instead represents knowledge that could be derived from a new set of games. Note that this could be an identity transformation if the target and input games are sufficiently similar (e.g., extracted jumping mechanic information shared between platforming games.)

\end{itemize}

PCG via Knowledge Transformation means casting the generative process as one of deriving, transforming, and applying knowledge from one (set of) games to another (set of) games. More formally:
\begin{gather}
    \mathbb{K}_1 = D_1(\mathbb{G}_1) \\
    T(\mathbb{K}_1) = \mathbb{K}_2 \approx D_2(\mathbb{G}_2) \\
    \mathbb{G}_2 \not\subseteq \mathbb{G}_1
\end{gather}

\noindent An overview of the application of this PCG-KT framework is depicted in Figure \ref{fig:ktpcgExample}.

\section{Framework for Describing PCG via Knowledge Transformation}\label{sec:framework}

In this section, we propose a set of features and axes for describing approaches to PCG-KT. Using the definitions presented in the previous section, we can frame many existing PCG approaches as exhibiting a form of knowledge transformation. For clarity, we will discuss a higher level descriptive category and then its underlying features and possible axes.

\subsection{Knowledge Structure}

When discussing and describing these transformation approaches, it is important to consider the \textit{structure} of the knowledge being transformed. Specifically, we consider structure at each of the three different stages of the transformation process; the raw input, the extracted input, and the transformed output.

\subsubsection{Raw Knowledge}
This is the knowledge closest to the input domains, meant to represent the input game(s) or domain(s) themselves, and is denoted: $$\mathbb{G}_i$$
Its structure depends on the context of the transformation, especially the domain (i.e., which game(s)); and the knowledge to be extracted from that domain (e.g., level designs, story structures, sprite palettes, character designs, etc.). For example, in a lot of PCGML work, a tile-based representation of a level can be taken as the raw knowledge, as in Figure \ref{fig:ktpcgExample}.
This enables methods for learning about the level structure in discrete patterns (e.g., \cite{snodgrass2016learning,summerville2016super}).
Alternatively, if the process started from images of input levels and  converted them into some ML-parsable representation like tiles, the input images would be considered the raw knowledge.
Other raw input formats such as segmented videos of playtraces enabled other methods to learn about gameplay relationships and object placements in the levels~\cite{guzdial2016learning}.
This is relevant to knowledge transformation as all later steps, including the possible output of the transformation, will depend upon this knowledge.
Thus, different raw knowledge formats have different affordances for a PCG-KT process. 
For example, the tile-based raw knowledge would not support working with portals as it does not include them.

\subsubsection{Extracted Knowledge}
This is the knowledge that is immediately extracted from the raw input knowledge prior to performing transformation, and is denoted:
$$\mathbb{K}_i = D_i(\mathbb{G}_i)$$

\noindent Its structure depends on the raw input knowledge, the methods used for extracting the knowledge, and the type of transformations to be performed. Such knowledge can be categorized based on whether it is direct or latent. 
Closer to the direct end, we could imagine a multi-layer~\cite{snodgrass2020multi} or hierarchical~\cite{snodgrass2015hierarchical} representation of the levels annotated with level features. Somewhere in the middle, we could have a learned probability table of tile distributions \cite{snodgrass2016learning}. At the latent end, we could have the weights of a neural network trained on those levels \cite{summerville2016super} or the latent vector representations of the levels \cite{sarkar2019blending,sarkar2020exploring}.

\subsubsection{Transformed Knowledge}
Lastly, we have the knowledge after the transformation has been performed, which is denoted:
$$T(\mathbb{K}_i) = \mathbb{K}_j \approx D_j(\mathbb{G}_j)$$

\noindent The structure of this knowledge can be described similarly as the structure of the knowledge prior to transformation, and so we will not repeat the same descriptions. It is important to note however that while many existing approaches use the same structure for the extracted and transformed knowledge, this is not required. For example, we can conceive of a method that takes a set of learned probability tables that model the level design, and outputs a set of latent vectors meant to blend those models together, or that takes a set of neural network weights from two models trained to produce different kinds of game content, and outputs content for a third, unseen game. 

\subsection{Derivation Function}
The derivation function is the process by which the extracted knowledge is derived from the raw knowledge described above. In the knowledge extraction equation $\mathbb{K}_i = D_i(\mathbb{G}_i)$, the derivation function is $D_i$. There are a variety of ways to construct the derivation function. We describe three broad categories ranging from hand-authored to hybrid to automated, representing a sliding scale where the function can be either more or less hand-authored (or automated), with fully manual and fully automated lying on the extreme ends of the spectrum.

\subsubsection{Hand-Authored}
Hand-authored derivation functions rely on manual authorship and domain expertise to pull the extracted knowledge from the raw knowledge. 
An example of such a function would involve a human domain expert analyzing a set of input levels and then devising a set of design rules for the given game (e.g., holes can only be $4$ tiles wide; cave levels can only have specific enemies, etc.). This extracted knowledge can then be used in the transformation. Another example could be manually tagging and labelling structures or patterns within content (levels, stories, quests, etc.).

\subsubsection{Automated}
At the other end of the spectrum, automated derivation functions rely on automated rules, heuristics or machine learned information in order to pull the extracted knowledge from the raw knowledge. For example, Summerville and Mateas~\cite{summerville2016super} and Snodgrass and Onta{\~n}{\'o}n~\cite{snodgrass2016learning} use backpropagation and occurrence counting, respectively, as the derivation function for pulling out the extracted knowledge. In those examples, the extracted knowledge is the trained model, be it a conditional probability distribution or an LSTM. If we think of the domain of stories or quests, we can imagine an approach that pulls out word embeddings from the narrative text to use as the extracted knowledge. The derivation function here could be a simple word clustering method, or a neural network approach that returns latent vector encodings.

\subsubsection{Hybrid}
These leverage both automated methods and human expertise. Most approaches in practice are likely to fall under this category, as even automated approaches may rely on domain expertise for setting up rules, training schema, labels, etc. Consider the automated approach of extracting a conditional distribution of tile types to model design knowledge from a game level. We can view this as a hybrid approach when the user is defining the neighborhood of tiles the model should be looking at for extracting the distribution. Alternatively, if we consider the hand-authored approach of tagging structures or patterns in stories, then the initial output of this approach could be used as the starting point for training a more automated approach to tag new examples. 

\subsection{Transformation Context}
The \textit{context} of the transformation or how the transformation knowledge is going to be used is another important factor to consider.
For describing the context, we propose the following axes, though these are not meant to be exhaustive.

\subsubsection{Novelty}
This describes if the transformation is intended to approximate an existing domain or explore new domains. For example, on the approximation end of the axis, Snodgrass and Onta{\~n}{\'o}n~\cite{snodgrass2016approach} try to learn a mapping of level representations from one game to another existing game, where some prior knowledge is known. On the other end, Guzdial and Riedl~\cite{guzdial2020conceptual} present an exploration approach where knowledge extracted from a set of input games is transformed and recombined to create entirely new games. Approaches can be placed along this spectrum based on how much prior knowledge of the approximated domain is known, and the extent to which the explored domains are different from the input domains.

\subsubsection{Usage}
This aspect of the context relates to the use of the approach and is less of an axis and more a categorical description of where the approach sits in a user context. For example, is the approach being used in a prototyping setting to explore new game ideas, a research environment to test the underlying methods, or to produce a complete new game for player consumption? Knowing this aspect of the context can directly influence what kinds of knowledge can be used and what the transformation function can look like.
Usage in this case is distinct from novelty, as for the same usage or goal, we might have different values of novelty. 
When prototyping new game mechanics, we might want very novel mechanics (i.e. distinct from all prior mechanics our system has seen) or we might more about how well the mechanics fit with other aspects of the design (e.g., narrative, aesthetics, etc.)

\subsection{Transformation Properties}
This is a broad category that encompasses many aspects of the transformation performed and is targeted towards describing the extent to which knowledge is transformed and the relationship between the transformed and input knowledge. 
This is distinct from just identifying input and output knowledge as in the definition, where we did not explicitly outline their possible relationships.
When describing the extent of the transformation, we need to consider the distance of the transformed knowledge from the input knowledge. We use the term \textit{distance} rather loosely. Instead of defining a specific distance metric, we discuss two types of distance depending on the structure of knowledge and the properties of interest. 

\subsubsection{Representational Distance}
This refers to the difference between the knowledge structure in the input and transformed knowledge. For example, the representational distance is quite low if interpolating between two latent vectors as all the knowledge has the same encoding. However, this distance is much higher when, for example, transforming a story graph into a set of three dimensional points in a game world. Notice, the representational distance for many approaches is fairly low as the knowledge structure is typically the same for the input and transformed knowledge (e.g., mapping one tile or sprite set to another, or modifying a learned distribution). 

\subsubsection{Content Distance}
This is the difference between the content of the input and transformed knowledge. 
For example, the content distance between sets of sprites extracted from two different Mario levels is much lower than the content distance between the sets of sprites extracted from a Mario level and a Mega Man level. 
If we consider two transformations, one that produces a new set of sprites by combining individual sprites from Mario and Mega Man, and one that creates a new set of sprites by randomly choosing a subset of sprites from Mario and Mega Man, then the content distance between the resulting blended sprite set and the input set would be larger than the distance between the resulting randomly joined sprite set and the input set. 
We can think of this as equivalent to the amount of shared information between the input and transformed knowledge, regardless of how that information is represented.
As an example, Sarkar et al.'s approach  \cite{sarkar2020exploring} of learning blended latent representations of several input games has a higher content distance than Snodgrass and Sarkar's approach \cite{snodgrass2020multi} that generates new levels by combining existing parts of existing game levels using binary space partitioning, which in turn has higher content distance than prior PCGML methods that operated on single-game domains.

Note that these two distances are independent of each other. Two sets of knowledge containing the same information but represented in different ways would have high representational distance but low content distance (e.g., representing level design knowledge for a game as tile-based representations of a set of levels vs. describing the levels as a set of rules for placing objects and enemies). 
Similarly, two sets of knowledge can represent different information in the same way. For example, two neural networks with the same architecture trained on two different games would result in knowledge sets with low representational distance, but high content distance as the weights of the networks would differ.

\subsection{Transformation Function}
The \textit{function} used to perform the transformation relates to all other aspects of PCG-KT since it relies on the \textit{context} of the transformation, needs the various knowledge \textit{structures} in order to perform the transformation and the \textit{properties} of the transformation depend upon this function. 
Here we break down the various ways of describing the function.

\begin{table*}[t!]
\centering
\begin{tabular}{l|l||p{3.5cm}|p{3.5cm}|p{3.5cm}||}

 & &  \hfil\textbf{\large{Game}}\hfil &  \hfil\textbf{\large{Conceptual}}\hfil & \hfil\textbf{\large{Domain}}\hfil \\

 & & \hfil\textbf{\large{Blending}}\hfil & \hfil\textbf{\large{Expansion}}\hfil & \hfil\textbf{\large{Transfer}}\hfil \\\hline\hline

\multirow{3}{*}{\parbox[c]{2cm}{\centering\large{Knowledge Structure}}}
&\textit{Raw Knowledge} & VGLC Levels & Gameplay Videos & VGLC Levels\\
&\textit{Extracted Knowledge} & VAE latent vectors & Game Graphs & MdMC Conditional Distribution  \\
&\textit{Transformed Knowledge}  & Blended VAE latent vectors via interpolation and labeling & Conceptual Expansion of Game Graphs & VGLC Levels Converted to Target Domain \bigskip\\\hline

\multirow{3}{*}{\rotatebox[origin=c]{0}{\parbox[c]{2.02cm}{\centering\large{Derivation Function}}}}
& \textit{Type} & Automated & Automated & Automated \\
& \textit{Description} & Training via Backpropagation & Rule Learning and Probability Estimation & Estimating Tile Distributions via Observation Counting\bigskip\\\hline

\multirow{3}{*}{\rotatebox[origin=c]{0}{\parbox[c]{2.02cm}{\centering\large{{Trans\-form Context}}}}}
&\textit{Novelty} & New & New & Transfer\\
&\textit{Usage} & Research Environment & Research Environment & Research Environment \bigskip\\ \hline

\multirow{3}{*}{\rotatebox[origin=c]{0}{\parbox[c]{2.02cm}{\centering\large{{Trans\-form Properties}}}}}
&\textit{Representation Distance}  & Low & Low & Low\\
&\textit{Content Distance} & High & High & Medium\bigskip\\\hline

\multirow{3}{*}{\rotatebox[origin=c]{0}{\parbox[c]{2.02cm}{\centering\large{{Trans\-form Function}}}}}
&\textit{Approach} & Interpolation/evolution of VAE latent vectors & Conceptual Expansion & Multi-stage Pipeline for Defining Mapping between Tiles\\
&\textit{Input Domain} &  Set of $N$ games & Set of $N$ games & Target Game and Input Game(s)\\
&\textit{Output Domain} & New game blending the $N$ games & New game represented as a combination of the $N$ games & Input Game(s) Transformed to Target Game Representation\\
&\textit{Cardinality} & Many-to-One, Many-to-Many & Many-to-One, Many-to-Many & One-to-one, Many-to-one\medskip\\
\end{tabular}
\caption{The PCG-KT framework applied to three techniques: game blending, conceptual expansion and domain transfer.}
\label{tbl:examples}
\end{table*}

\subsubsection{Approach}
This refers to the actual procedure or manner in which the transformation is being performed. This can be through updating a probability table \cite{snodgrass2016learning}, backpropagation through a neural network \cite{summerville2016super}, iteratively growing a graph from existing extracted graphs  \cite{guzdial2020conceptual}, interpolating between latent vectors \cite{sarkar2019blending,sarkar2020exploring}, or reducing detailed knowledge representations from various domains into a single uniform representation \cite{snodgrass2020multi,sarkar2020exploring}, to name a few. For this category, the function used can be nearly any procedure already used in the PCG, ML, creativity, design, or AI literature, including manual and hybrid approaches. 
Explicitly identifying the procedure where the transformation takes place can help to more clearly delineate the transformation approach as a whole. 

\subsubsection{Input/Output Domains}
The {input and output domains} of transformation help characterize the nature of the transformation process.
For example, in works centered around blending different games, the input domain is the set of games to be blended and the output domain is the final blended game. Thus, here the transformation process is one of transforming input knowledge extracted from the individual games into output knowledge recombining properties of the input games together. Similarly, in the case of domain transfer for level design, the input domain consists of a set of one or more input games and a target game and the output domain consists of levels of the input game(s) transformed to the target game representation.


\subsubsection{Cardinality}
This refers to the number of input sources and number of outputs. We consider four types of cardinality:
\begin{enumerate}
\item  \textit{One input domain to one output domain}: this is a standard knowledge transfer where knowledge from one domain is transformed to represent knowledge from another domain \cite{snodgrass2016approach}. 
\item \textit{Multiple inputs to one output}: this is typically a blending approach where the input domains are transformed and combined in some way to get a different output domain (either new or existing) as in \cite{snodgrass2020multi,sarkar2020exploring,guzdial2020conceptual}. 
\item \textit{Multiple inputs to multiple outputs}: this can also be blending or approximating, but likely in a more exploratory way where the input knowledge is transformed and recombined in various ways to explore different output domains. Examples of this are \cite{sarkar2020conditional,sarkar2021dungeon} where the input is a group of games but rather than produce a single blended output, different labels are used to produce different combinations of blends where each unique label configuration can be thought of as mapping the set of input games to a separate blended domain made up of a unique combination of the input games.
\item \textit{One input to multiple outputs}: this is also a more exploratory transformation where the knowledge from a single domain is transformed and possibly abstracted in order to explore multiple output domains.
\end{enumerate}
To the best of our knowledge, while there have been systems that combined multiple inputs into multiple outputs consecutively, there have not been any systems that have attempted to simultaneously generate multiple outputs dependent on one another (e.g., a series of games, a set of sprites, or a soundtrack). 

To illustrate the framework using concrete examples, Table \ref{tbl:examples} depicts the result of applying the PCG-KT framework to three separate applications, game blending~\cite{sarkar2019blending}, conceptual expansion~\cite{guzdial2020conceptual}, and domain transfer~\cite{snodgrass2016approach}, showing how the features and axes discussed above can be categorized in each case. 
The raw knowledge $\mathbb{G}_i$ consists of VGLC levels for game blending and domain transfer while consisting of gameplay videos in the case of conceptual expansion. In all three cases, the derivation function ($D_i(\mathbb{G}_i)$) involves a machine learning algorithm (such as a neural network, Bayes net, or Markov model) which is used to obtain the extracted knowledge ($\mathbb{K}_i$). This is in the form of latent vectors for game blending, game graphs for conceptual expansion, and conditional distribution tables for domain transfer. For the transformation function ($T(\mathbb{K}_i)$), conceptual blending employs latent vector interpolation, whereas conceptual expansion requires a multi-stage pipeline including defining tile mappings to obtain the final transformed knowledge $\mathbb{K}_j$. This consists of blended latent vectors for game blending, a recombined game graph representing a new game in the case of conceptual expansion, and levels in the target domain for domain transfer.
These applications leverage two broad categories of transformation functions---combinational creativity and transfer learning. These and the examples in Table \ref{tbl:examples} are discussed in more detail in Section \ref{sec:transformationfunctions}.

\section{Related Work}

In this section, we discuss related areas of prior work and compare them to the focus of this paper.

\subsection{Procedural Content Generation}

PCG-KT is a framework for thinking about a particular PCG process, thus all prior PCG work represents a related area. 

One clear area of related work is Procedural Content Generation via Machine Learning (PCGML) \cite{summerville2018procedural}, given that we identify PCG-KT as a solution to major problems with this approach. 
However, we believe PCG-KT also applies to cases where the input knowledge is hand-authored, as in the work of Permar and Magerko \cite{permar2013conceptual}.
Thus we identify PCG-KT as applying to PCGML, but not wholly subsumed by it.

Supervised PCGML attempts to take in knowledge in the form of existing content and transform it (via training an ML model and then sampling from that model) into additional content.
However, we do not consider this process to be PCG-KT in most cases due to a difference in goals.
For example, a PCGML approach might attempt to learn a model of level design that allows for the generation of content (e.g., new levels) that would fit some existing game (e.g., Super Mario Bros.)~\cite{summerville2016super,snodgrass2014experiments}. 
In this case, the goal is explicitly that the set of output levels should substantially resemble the original knowledge. 
This is typically done by learning a distribution based on the original, existing content and then sampling from that distribution.
It follows then that any output from this learned model represents the same knowledge from the original game $\mathbb{G}$.
While our definition of PCG-KT covers the case in which the input game matches the output target game, this represents a minimal or trivial transformation.

We also identify PCG-KT as distinct from most search-based PCG (SBPCG) approaches, such as genetic algorithms or evolutionary search~\cite{togelius2010search}. 
In these approaches, we can consider this search space as having been defined by the author's own game design knowledge.
Thus, it is derived from a theoretical set of games $\mathbb{G}$, in this case all of the games the author considers when they define the search space.
It then follows that any output from this search space represents knowledge from $\mathbb{G}$.
There is prior work where an entire playable game is produced via SBPCG~\cite{cook2016angelina}. 
However, even in these cases, that game already existed in the unchanged search space. 
Thus, the output games must necessarily exist in $\mathbb{G}$.
In other words, because the author's definition of a space of games included the output game, there is no transformation that has occurred and the knowledge remains unchanged.
Note however that search-based methods could be  classified as PCG-KT if the search is used to perform knowledge transformation. The appropriateness of applying PCG-KT as a lens is dependent on the purpose of a system, not the system's technical approach.

\subsection{Design Patterns}

In software engineering, the design structures that make software more flexible and reusable are known as design patterns~\cite{buschmann2008pattern}.
The use of design patterns spans several disciplines and in general, a design pattern describes a recurrent problem in an environment and maps it to a core of solutions that can be reapplied and adapted without needing to be reinvented from scratch~\cite{alexander1977pattern}. 
Because the design of a game involves activities that are not related entirely to problem solving and engineering skills, the concept of game design patterns was introduced to address game mechanics, game design, and game art~\cite{gamedesignpatt, kreimeier2002case}. 
Prior works have used game design patterns to analyze level designs \cite{dahlskog2012patterns,10.1145/1822348.1822359,khalifa2019level} and interactive narrative \cite{10.1145/3102071.3106366}.
Björk and Holopainen \cite{gamedesignpatt} defined game design patterns as pieces of game design knowledge which could be identified when they occurred in multiple games (e.g., a companion character or collectible). 
There has been work that takes these design patterns and attempts to apply them to new games~\cite{hullett2013cause}.
However, we note that in these cases there is no effort made to transform these patterns outside of their initial identification. 
Thus a single design pattern applied to some new game is understood to be the same pattern as that in the original games.

\begin{table*}[t!]
\centering
\begin{tabular}{|cc|l|c|c|}
\hline
\multicolumn{2}{|c|}{Transformation Function} & \multicolumn{1}{c|}{\multirow{2}{*}{Papers}} & \multirow{2}{*}{Method} & \multirow{2}{*}{\begin{tabular}[c]{@{}c@{}}Generated \\ Content\end{tabular}} \\ \cline{1-2}
\multicolumn{1}{|c|}{\begin{tabular}[c]{@{}c@{}}Main \\ Category\end{tabular}} & \begin{tabular}[c]{@{}c@{}}Secondary \\ Category\end{tabular} & \multicolumn{1}{c|}{} &  &  \\ \hline
\multicolumn{1}{|c|}{\multirow{9}{*}{\begin{tabular}[c]{@{}c@{}}Combinational\\ Creativity\end{tabular}}} & \multirow{7}{*}{\begin{tabular}[c]{@{}c@{}}Conceptual\\ Blending\end{tabular}} & \cite{guzdial2016learning} & Hierarchical Bayes Network & Levels \\ \cline{3-5} 
\multicolumn{1}{|c|}{} &  & \cite{sarkar2019blending, sarkar2020exploring} & Variational Autoencoder (VAE) & Levels \\ \cline{3-5} 
\multicolumn{1}{|c|}{} &  & \cite{summerville2017mechanics}& Variational Autoencoder (VAE) & Jumps \\ \cline{3-5} 
\multicolumn{1}{|c|}{} &  & \cite{sarkar2020conditional, sarkar2021dungeon} & Conditional VAE & Levels \\ \cline{3-5} 
\multicolumn{1}{|c|}{} &  & \cite{nelson2007towards} & Graph Structure Knowledge Base & Games \\ \cline{3-5} 
\multicolumn{1}{|c|}{} &  & \cite{gow2015towards} & Manual Conceptual Blending & Games \\ \cline{2-5} 
\multicolumn{1}{|c|}{} & \multirow{2}{*}{\begin{tabular}[c]{@{}c@{}}Conceptual\\ Expansion\end{tabular}} & \cite{guzdial2018combinatorial}  & Bayesian Networks & Levels \\ \cline{3-5} 
\multicolumn{1}{|c|}{} &  & \cite{guzdial2018automated, guzdial2020conceptual} & Probabilistic Graphical Models & Games \\ \hline
\multicolumn{1}{|c|}{\multirow{6}{*}{\begin{tabular}[c]{@{}c@{}}Transfer\\ Learning\end{tabular}}} & \multirow{2}{*}{\begin{tabular}[c]{@{}c@{}}Domain\\ Adaptation\end{tabular}} &  \cite{snodgrass2016approach} & Multidimensional Markov Chains & Levels \\ \cline{3-5} 
\multicolumn{1}{|c|}{} &  & \cite{snodgrass2019levels, snodgrass2020multi} & VAE, Binary Space Partitioning & Levels \\ \cline{2-5} 
\multicolumn{1}{|c|}{} & \multirow{2}{*}{Style Transfer} & \multirow{2}{*}{} \hspace{-0.7em} \cite{sarkar2022tile} & \multirow{2}{*}{\begin{tabular}[c]{@{}c@{}}Tile-to-Affordance Mapping\\ Autoencoder/Markov Model\end{tabular}} & \multirow{2}{*}{Levels} \\
\multicolumn{1}{|c|}{} &  &  &  &  \\ \cline{2-5} 
\multicolumn{1}{|c|}{} & Design Patterns & \cite{beaupre2018design} & Case-based Reasoning & Levels \\ \cline{2-5} 
\multicolumn{1}{|c|}{} & \begin{tabular}[c]{@{}c@{}}Data Mining \\ Association\end{tabular} & \cite{machado2019pitako} & Apriori Algorithm & \begin{tabular}[c]{@{}c@{}}Game \\ Rules\end{tabular} \\ \hline
\end{tabular}
\caption{Summary of the PCG-KT works discussed in Section V.}
\label{tab:category}
\end{table*}

\subsection{Transfer Learning and Domain Adaptation}\label{sec:RW-DA}

In transforming knowledge from one domain to another, PCG-KT touches upon transfer learning and domain adaptation and in later sections, we will see examples of PCG-KT work that leverages these techniques. Zhuang et al. \cite{zhuang2020comprehensive} define transfer learning methods as those that attempt to leverage knowledge learned for a source task and use this source knowledge to benefit the learning of knowledge for a target task. Domain adaptation is defined by Pan et al. \cite{pan2010domain} as the method of solving a learning problem in a target domain by using training data in a different but related source domain. 
In games, other than the works discussed in this paper, such methods have primarily been used for transferring the behavior of agents from one environment to another. 
Chaplot et al. \cite{chaplot2016transfer} used deep Q-networks to transfer agent navigation behavior from seen to unseen maps in the VizDoom environment. 
Note that this does not fall under PCGKT since the knowledge transfer happens within the same game. Consequently, no content is generated for any new game.
Rusu et al. \cite{rusu2016progressive} introduced progressive networks for transfer learning agent behavior across different Atari games while Kansky et al.~\cite{kansky2017schema} used schema networks to learn basic physics models of entities, which were then used in novel configurations of Breakout not seen during training. 
Melnik et al.~\cite{melnik2018functional} learned modules such as object detection and trajectory based physics to transfer an agent policy from Pong to Breakout.
Relatedly, several approaches have been attempted to ease the ability to transfer knowledge learned in one game and apply it to another.  
Mittel and Munukutla~\cite{mittelAndMunukutla} and  Gamrian and Goldberg~\cite{gamriantransfer} handle this task by disentangling the visual aspects from the control aspects of a learned agent. 
\cite{mittelAndMunukutla} transfer agent policies from Atari Pong to Breakout while \cite{gamriantransfer} perform transfer learning to build agents capable of playing Breakout and Road Fighter.
However, note that in each of \cite{rusu2016progressive}-\cite{gamriantransfer}, while some knowledge concerning agent behavior is being transferred from one game to another,
no new content is created as a result of the transfer, thus failing to fall under the PCGKT definition.
In \cite{kim2020learning}, Kim et al. introduced GameGAN, a neural network that can learn a model of a game by separately modeling the static and dynamic elements. 
This allows swapping the static and dynamic elements of different games to perform a sort of transformation.
However, the transformation is minimal, closer in nature to swapping out sprites or background images than a transformation of game structure or mechanics.

\subsection{Computational Creativity}

Computational Creativity (CC) is the umbrella term for efforts to reproduce human creativity in computers \cite{colton2012computational}. 
Since we focus on how knowledge related to PCG can be transformed to produce novel output in a particular creative process, it relates to CC. 
CC has long been linked to games \cite{liapis2014computational}, however the majority of this work has focused on ``exploratory'' creativity, which is strongly associated with SBPCG discussed previously. 
A more related type of creativity to PCG-KT is ``transformational'' creativity~\cite{boden1996creativity}, where a conceptual space, which can be thought of as analogous to a search space or distribution, is transformed via an alteration in the variables representing the space.
This has strong parallels to our PCG-KT definition, though transformational creativity is typically much more loosely defined. 
Boden defined a third type of creativity: combinational creativity~\cite{boden1996creativity}.
This type of creativity refers to the combination of concepts, and has inspired a large number of computational processes to attempt to replicate it, with the most famous being conceptual blending~\cite{fauconnier1998conceptual}.
This approach is known in the psychology field as conceptual combination~\cite{Gagn1997}, and has been studied for decades as a ubiquitous cognitive process. 
One of the two categories we introduce for transformation functions is based on combinational creativity. 
There are other CC methods that have some overlap with PCG-KT. 
For example, novelty search \cite{lehman2011novelty} and surprise search~\cite{gravina2016surprise}, which attempt to locate members of a search space or environment that are novel and surprising respectively. 
While this might seem to be equivalent to PCG-KT, because the search space remains unchanged (the knowledge it represents remains untransformed), we do not consider it to be the same process. 
However, we could imagine employing approaches like novelty or surprise search towards PCG-KT, seeking novel or surprising transformations, and thus we consider these and other CC methods to be complementary to those discussed in this paper.

\begin{figure}[t]
\centering
\setlength\tabcolsep{1pt}
\begin{tabular}{cccccc}
\includegraphics[width=0.075\textwidth]{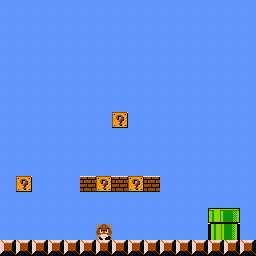}
\includegraphics[width=0.075\textwidth]{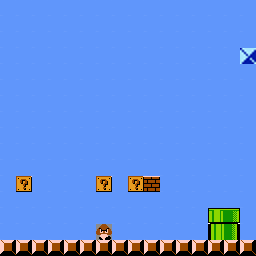}
\includegraphics[width=0.075\textwidth]{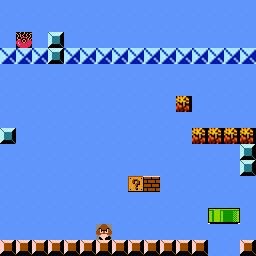}
\includegraphics[width=0.075\textwidth]{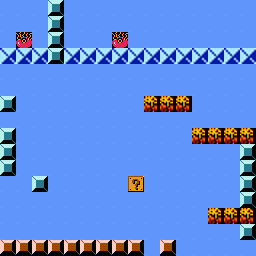}
\includegraphics[width=0.075\textwidth]{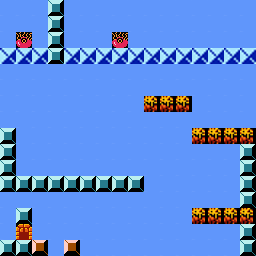}
\includegraphics[width=0.075\textwidth]{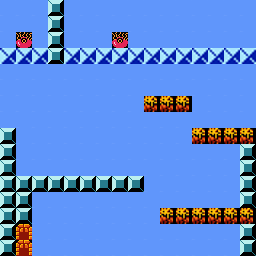}
\end{tabular}
\caption{\label{fig:sarkar2019} Example of blending between \textit{Super Mario Bros.} and \textit{Kid Icarus}. Reproduced with permission from \cite{sarkar2019blending}.}
\end{figure}

\begin{figure*}[t]
\centering
\setlength\tabcolsep{1pt}
\begin{tabular}{ccccc}
\includegraphics[width=0.19\linewidth]{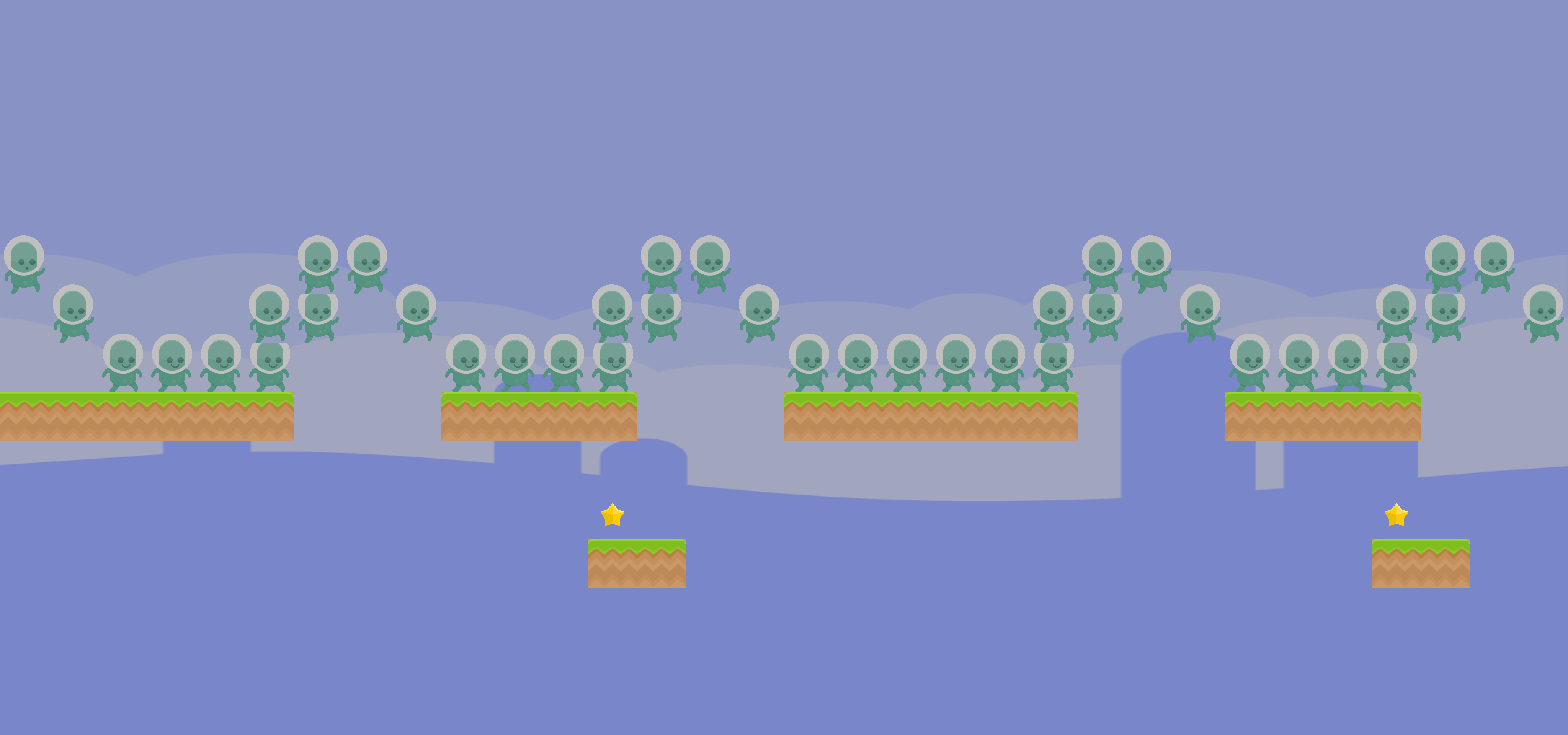}
\includegraphics[width=0.19\linewidth]{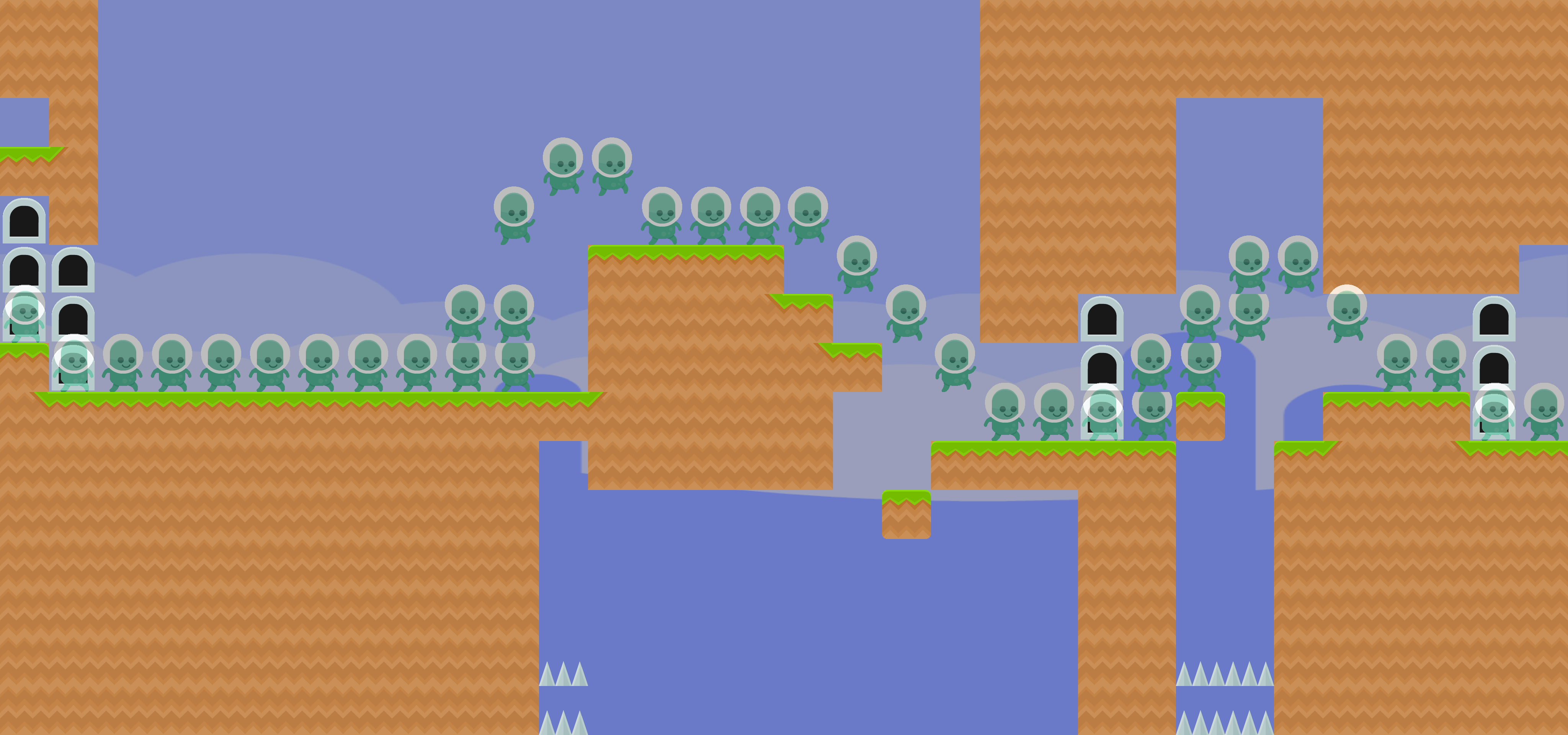}
\includegraphics[width=0.19\linewidth]{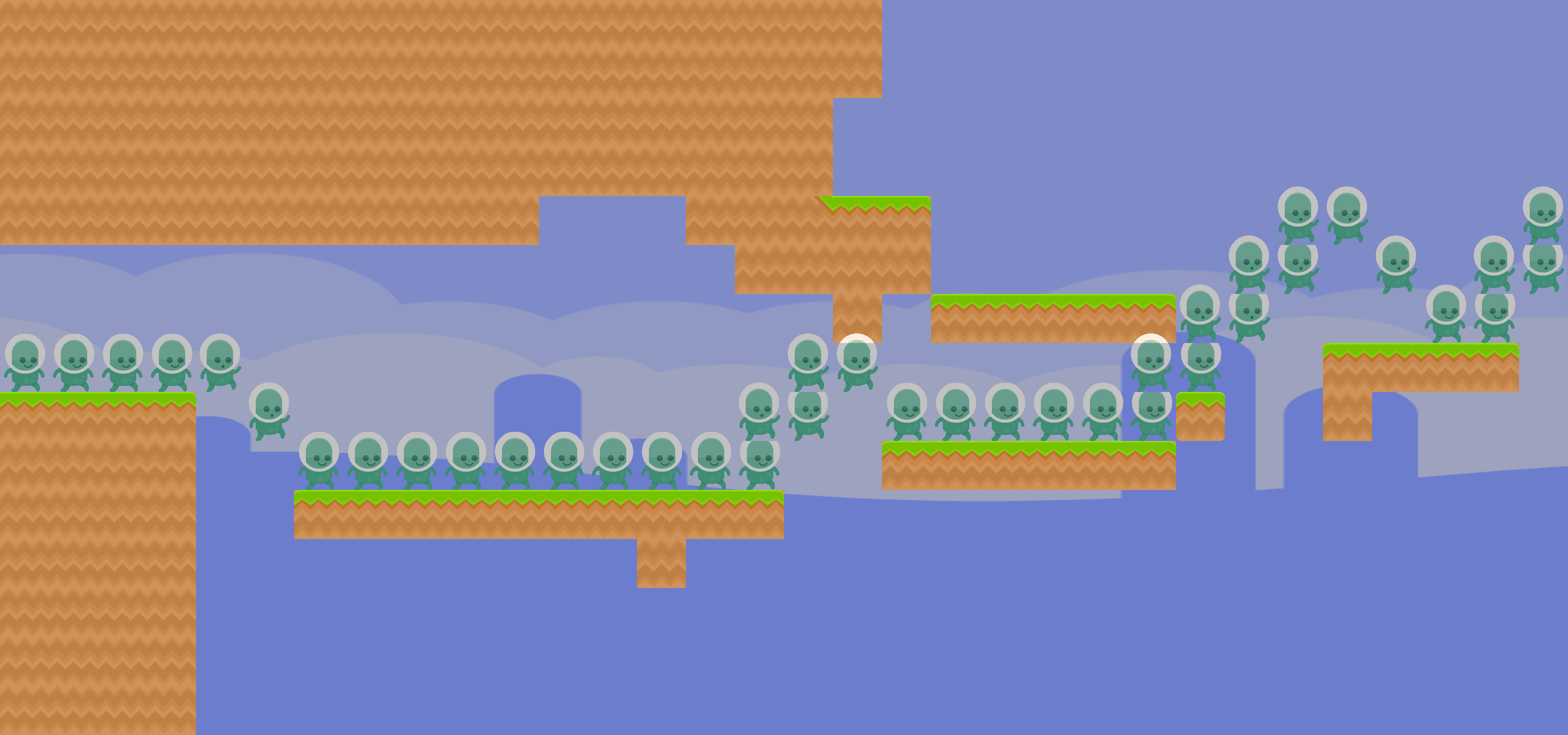}
\includegraphics[width=0.19\linewidth]{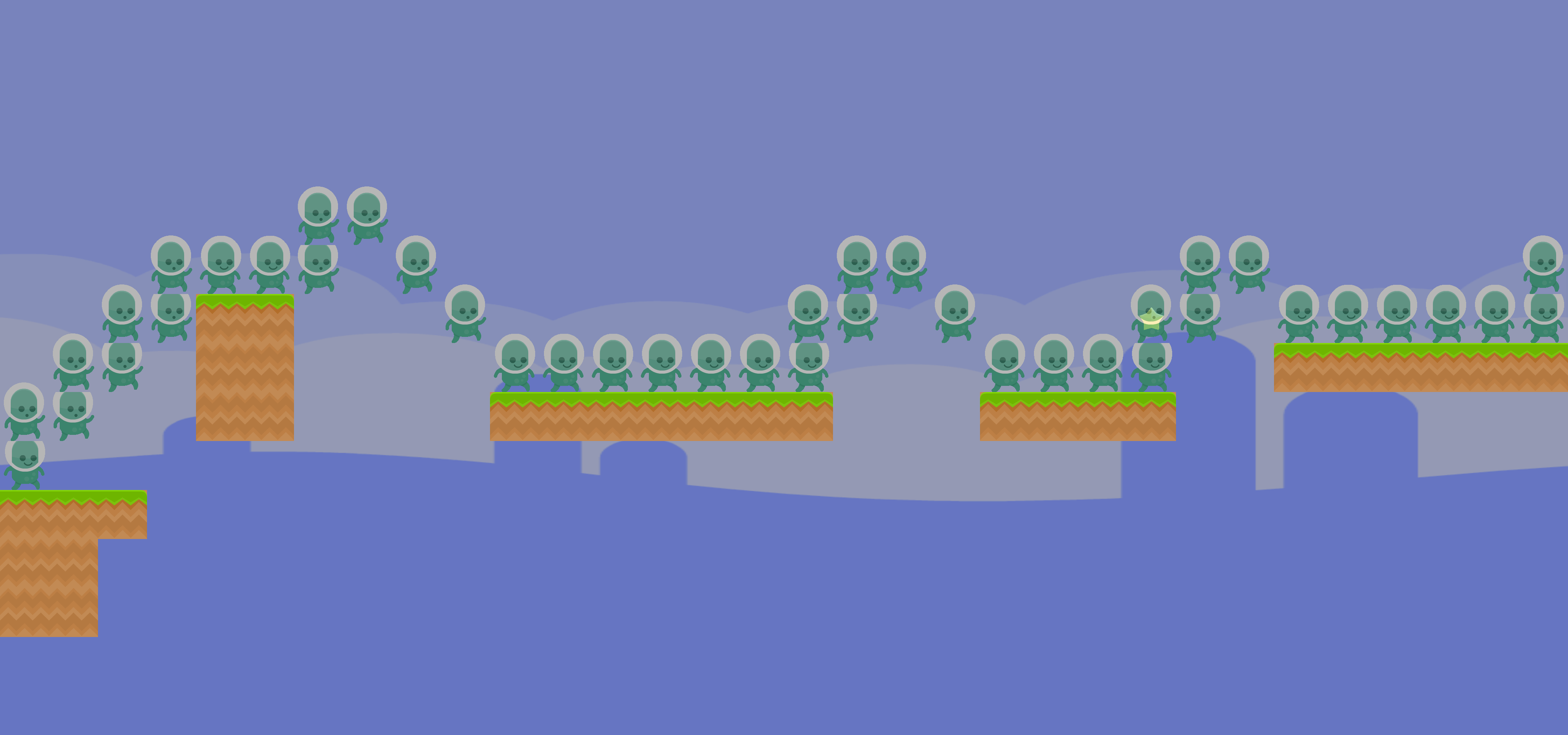}
\includegraphics[width=0.19\linewidth]{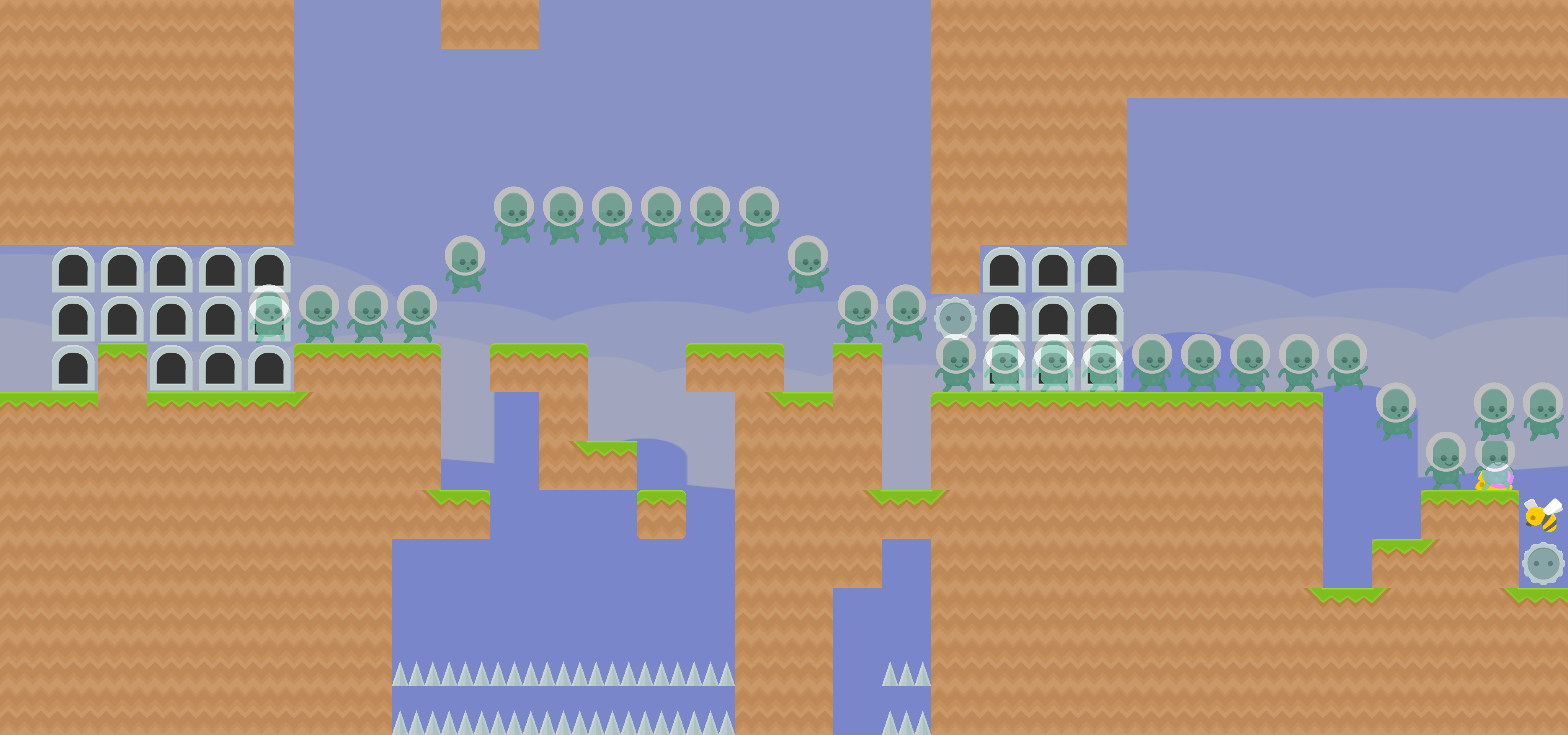}
\end{tabular}
\caption{\label{fig:blap} Example interpolation between \textit{Mega Man} and \textit{Metroid}. Reproduced with permission from \cite{sarkar2020exploring}}
\end{figure*}

\begin{figure*}[tb]
  \centering
  \includegraphics[width=0.975\linewidth]{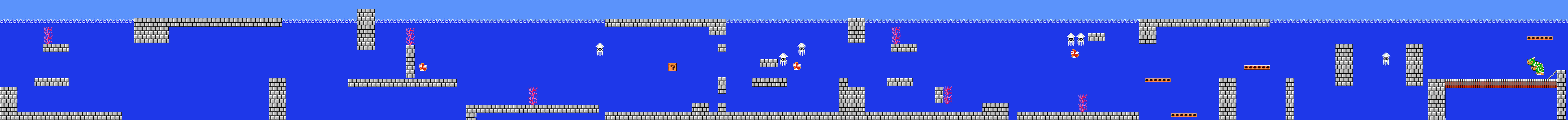}
  \caption{Example output of a blended level design model.}
  \label{fig:guzdial2016learning}
\end{figure*}

\begin{figure}[t]
\centering
\includegraphics[width=0.9\linewidth]{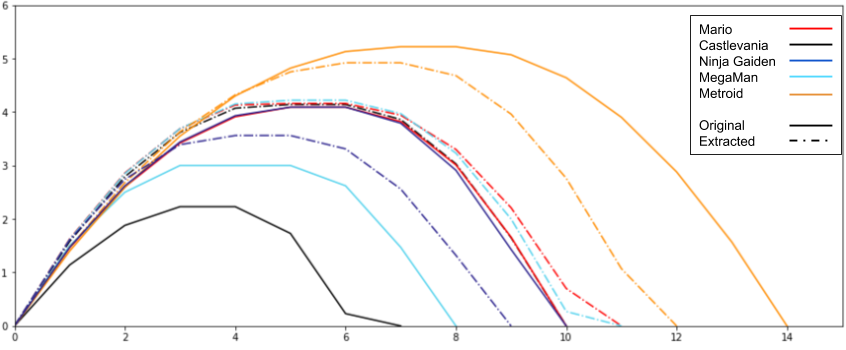}
\caption{Original game jumps (solid) vs generated jumps (dash-dotted) extracted from blended domains. Reproduced with permission from \cite{summerville2020extracting}.}
\label{fig:TrueVsGen}
\end{figure}

\begin{figure*}[t]
\centering
\setlength\tabcolsep{1pt}
\setlength{\fboxsep}{0pt}
\setlength{\fboxrule}{1.75pt}
\begin{tabular}{ccccccccc}
\raisebox{12pt}{\rotatebox{90}{\scriptsize{SMB}}}
\includegraphics[width=0.15\columnwidth]{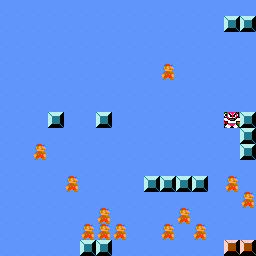} &
\includegraphics[width=0.15\columnwidth]{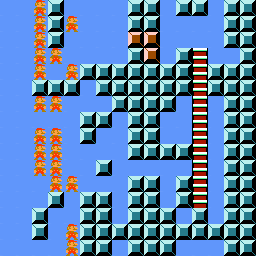} &
\includegraphics[width=0.15\columnwidth]{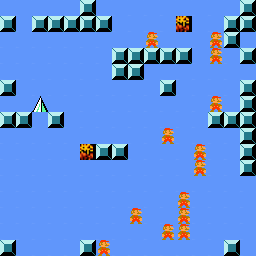} &
\includegraphics[width=0.15\columnwidth]{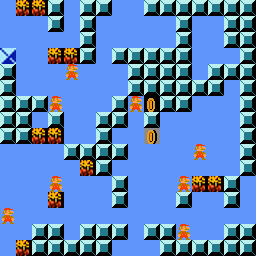} &
\fbox{\includegraphics[width=0.15\columnwidth]{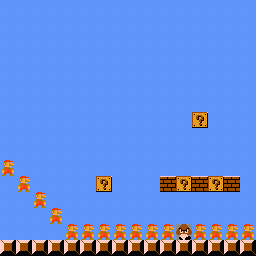}} &
\includegraphics[width=0.15\columnwidth]{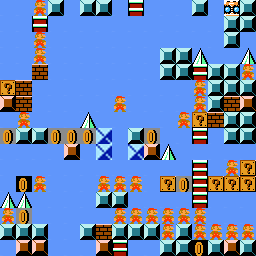} &
\includegraphics[width=0.15\columnwidth]{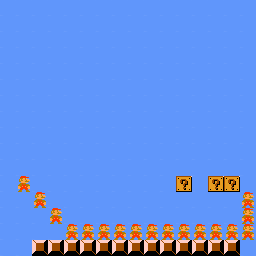} &
\includegraphics[width=0.15\columnwidth]{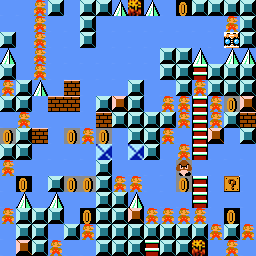} &
\\
\raisebox{15pt}{\rotatebox{90}{\scriptsize{KI}}}
\includegraphics[width=0.15\columnwidth]{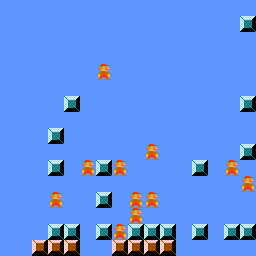} &
\includegraphics[width=0.15\columnwidth]{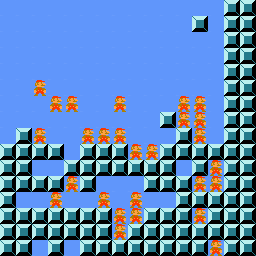} &
\fbox{\includegraphics[width=0.15\columnwidth]{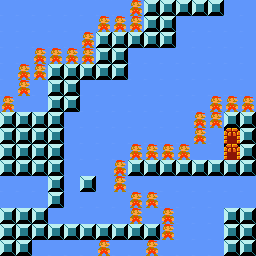}} &
\includegraphics[width=0.15\columnwidth]{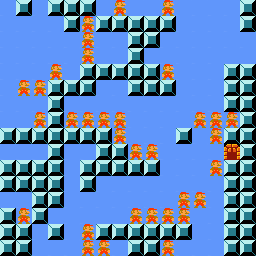} &
\includegraphics[width=0.15\columnwidth]{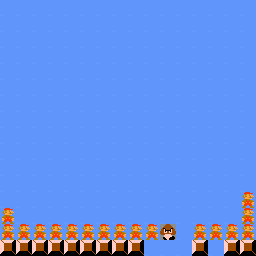} &
\includegraphics[width=0.15\columnwidth]{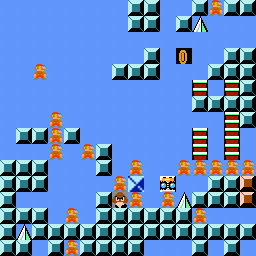} &
\includegraphics[width=0.15\columnwidth]{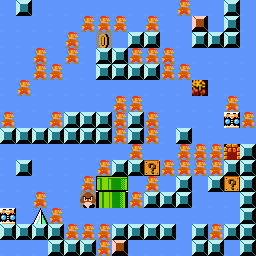} &
\includegraphics[width=0.15\columnwidth]{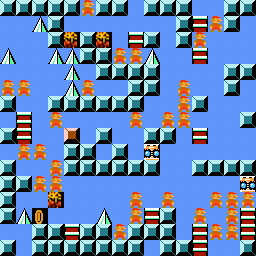} &
\\
\raisebox{15pt}{\rotatebox{90}{\scriptsize{MM}}}
\includegraphics[width=0.15\columnwidth]{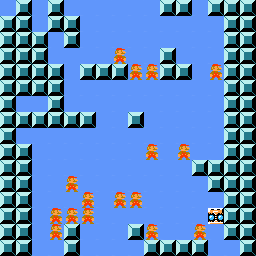} &
\fbox{\includegraphics[width=0.15\columnwidth]{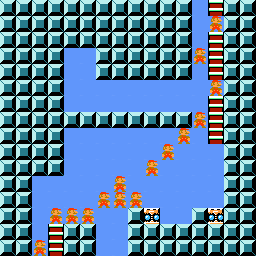}} &
\includegraphics[width=0.15\columnwidth]{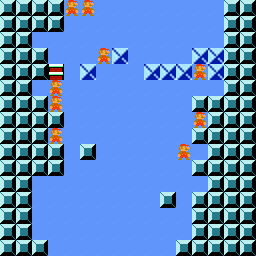} &
\includegraphics[width=0.15\columnwidth]{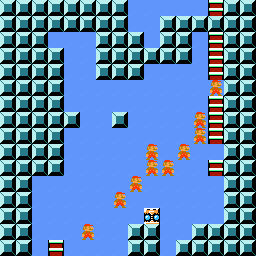} &
\includegraphics[width=0.15\columnwidth]{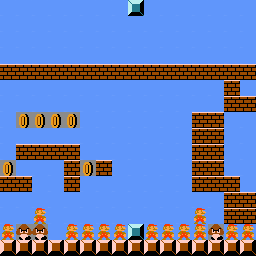} &
\includegraphics[width=0.15\columnwidth]{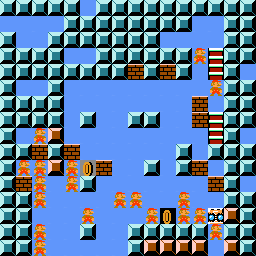} &
\includegraphics[width=0.15\columnwidth]{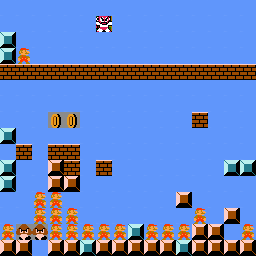} &
\includegraphics[width=0.15\columnwidth]{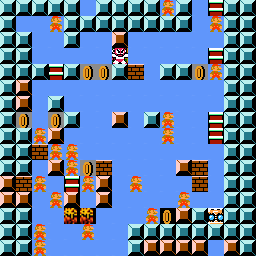} &
\\
000 & 001 & 010 & 011 & 100 & 101 & 110 & 111
\end{tabular}
\caption{\label{fig:cvae} Segments generated by conditioning on blend labels (shown below the 3rd row), using an original segment from \textit{Super Mario Bros.} (SMB) (1st row), \textit{Kid Icarus} (KI) (2nd) and \textit{Mega Man} (MM) (3rd). First, second and third elements of the label correspond to SMB, KI and MM respectively. Bordered segments are originals. Reproduced with permission from \cite{sarkar2020conditional}}
\end{figure*}

\begin{figure}[tb]
  \centering
  \includegraphics[width=\linewidth]{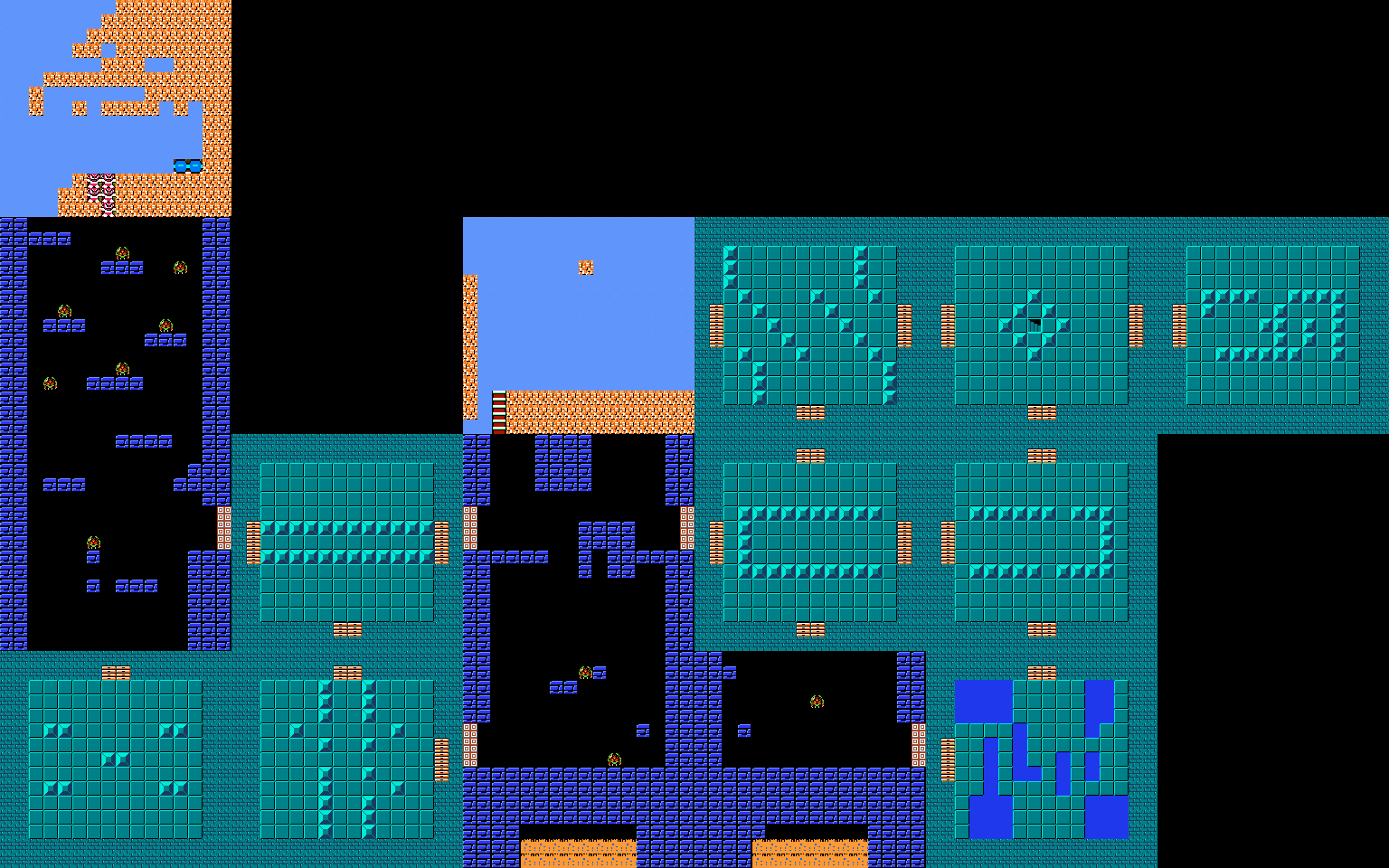}
  \caption{Example level blending \textit{The Legend of Zelda}, \textit{Metroid} and \textit{Mega Man}. Reproduced with permission from \cite{sarkar2021dungeon}}
  \label{fig:sarkar2021dungeon}
\end{figure}

\section{Categories of Transformation Functions}\label{sec:transformationfunctions}

In this section, we survey existing literature that fits under the banner of PCG-KT. 
While we introduced a definition of PCG-KT earlier and several aspects to consider when designing a PCG-KT system, we anticipate that the primary differentiator of PCG-KT systems is their transformation functions: the process that actually changes the knowledge.
Thus, we identify two broad categories of PCG-KT transformation functions: (1) combinational creativity and (2) transfer learning. 
In the first, the transformation is based on combining existing PCG knowledge to produce other PCG knowledge capable of generating novel content. 
In the second, knowledge is transformed via the use of transfer learning~\cite{zhuang2020comprehensive}, taking knowledge from a source domain and altering it to better fit some target domain.
These roughly correspond to many-to-many and many-to-one cardinality for combinational creativity and one-to-one cardinality for transfer learning.
We discuss both categories and cover any subcategories and particular recent approaches that fall into each category. To help situate each discussed work in the PCG-KT framework, we highlight the form of the input knowledge and that of the output knowledge obtained after applying the transformation process.
Here, input and output knowledge refer to the input extracted knowledge and output transformed knowledge of the transformation function ($T(\mathbb{K}_i) = \mathbb{K}_j$) described previously.
An overview of the discussed works is given in Table \ref{tab:category}.

\subsection{Combinational Creativity}

Combinational creativity is a type of creativity that deals with the creation of new artifacts, ideas and/or knowledge via recombining existing knowledge in novel, unfamiliar ways \cite{boden2009computer}.
Guzdial and Riedl drew on combinational creativity in an early argument as to how to solve the problem of producing novel content with supervised PCGML~\cite{guzdial2018combinatorial}.
There are a great number of approaches to computationally represent combinational creativity.
These approaches include amalgamation~\cite{ontanon2010amalgams} and compositional adaption \cite{badie2017compositional}, but these approaches have largely not been applied to PCG problems. 
Thus we focus on two specific combinational creativity approaches: conceptual blending and conceptual expansion. 

\subsubsection{Conceptual Blending}
Conceptual blending is a combinational creativity technique focusing on producing novel concepts by combining or blending elements from existing concepts. Fauconnier and Turner \cite{fauconnier1998conceptual} formalized conceptual blending in their `four space' theory which describes a concept blend as comprising of four spaces - two \textit{input} spaces consisting of the concepts to be blended or combined, a \textit{generic} space into which the two input concepts are projected in order to find equivalence points amenable for blending, and a \textit{blend} space into which these points of equivalence are projected and from which novel concepts and patterns are evolved. In the following sections, we review prior work that utilizes conceptual blending and how they can be specified under the PCG-KT framework.

In \cite{guzdial2016learning}, Guzdial and Riedl employed conceptual blending on learned level design models. 
Here, the input knowledge consisted of level design models represented as hierarchical Bayesian networks, thus they were graph-like in nature. 
This made it simple to adapt the traditional conceptual blending algorithm to these models. 
The output knowledge was produced by blending the models to produce new level design models that could in turn generate new kinds of levels. 
For example, combining two generative level design models that produced underwater and castle Mario levels to create a model that could generate underwater-castle levels as in Figure \ref{fig:guzdial2016learning}.

In \cite{nelson2007towards}, Nelson and Mateas outlined an approach to recombine game mechanics in order to generate new games. The generation process involved taking a verb and/or noun from users to describe a theme and generating an appropriate game in response, using a set of pre-defined abstract game templates, each of which could be implemented using different stock sets of mechanics. The input knowledge here thus consisted of the game templates and mechanic sets selected for combination based on user-supplied verbs and nouns and the output knowledge was the resulting generated game produced via combining the selected template and mechanics.

In \cite{gow2015towards}, Gow and Corneli proposed generating entire games by blending elements from existing ones. In their framework, the input knowledge comprised of game specifications in the Video Game Description Language (VGDL) \cite{schaul2014extensible}. As an example, they manually applied conceptual blending as the transformation function on the VGDL specifications of the games \textit{Frogger} \cite{frogger} and \textit{The Legend of Zelda} \cite{zelda}. This produced as output knowledge the VGDL specification for \textit{Frolda}, a new game combining \textit{Frogger's} mechanic of crossing a road full of hazards with \textit{Zelda's} lock-and-key mechanic.

This framework has been operationalized through the use of ML in works by Sarkar and Cooper \cite{sarkar2019blending} and Sarkar et al. \cite{sarkar2020exploring}. Both use variational autoencoders (VAEs) \cite{kingma2013autoencoding} which consist of encoder-decoder networks that learn continuous, latent representations of the input data. By training on levels from multiple games, the learned VAE latent space spans all the games used in training and enables sampling blended levels. In these works, the input knowledge is comprised of the learned latent representations extracted from the original game levels (taken from \textit{Super Mario Bros.} \cite{supermario:nes}, \textit{Kid Icarus} \cite{kidicarus}, \textit{Metroid} \cite{metroid}, \textit{Mega Man} \cite{megaman}, \textit{Castlevania} \cite{castlevania} and \textit{Ninja Gaiden} \cite{gaiden}) along with the mapping between levels and latent vectors learned by the VAE encoder and decoder. The output knowledge is composed of the latent representations of blended levels obtained by applying conceptual blending via interpolation between latent vectors or by evolving such vectors. Example output levels are shown in Figure \ref{fig:sarkar2019} and Figure \ref{fig:blap} respectively. In follow-up work, Summerville et al. \cite{summerville2020extracting} extracted jump mechanics using paths in the generated blended levels. Example extracted jump arcs are shown in Figure \ref{fig:TrueVsGen}. Here, the output knowledge additionally included the blended jump mechanics extracted from the blended levels.

In follow-up work, Sarkar et al. \cite{sarkar2020conditional} and Sarkar and Cooper \cite{sarkar2021dungeon} used conditional VAEs which are VAEs that additionally use labels to control properties of generated, blended levels such as which combination of games to blend as well as orientation. In terms of the PCG-KT framework, an interesting distinction in these CVAE-based works is that in producing multiple blended domains as outputs, they exhibit a many-to-many cardinality as opposed to the many-to-one cardinality of the VAE-based works since the latter only produce a single blended domain as output. Example levels from \cite{sarkar2020conditional} are shown in Figure \ref{fig:cvae}. An additional point to note is that by enabling control over orientation, the approach in \cite{sarkar2021dungeon} enables producing levels combining dungeon rooms with platformer level segments (e.g. Figure \ref{fig:sarkar2021dungeon}). Thus this approach has an even higher content distance between input and transformed knowledge than prior methods which blend only platformers. Note that in these conditional approaches, the input and output knowledge is similar to that of VAE-based methods, but here the transformation function of conceptual blending is performed via label manipulation.

\subsubsection{Conceptual Expansion}

Guzdial and Riedl introduced conceptual expansion as an alternative to traditional, non-latent conceptual blending meant to allow for the blending of machine learned knowledge from more than two sources \cite{guzdial2018combinatorial}. 
The intuition is that it breaks the combinational creativity process into a series of functions of the form $CE^X(\mathrm{A},F) = \alpha_0*f_0+\alpha_1*f_1...\alpha_n*f_n$. 
Here, $F$ is a set of existing, machine-learned features, $\mathrm{A}$ is a set of filters that describe what to take from each feature and $X$ is a goal. 
$X$ may be an explicit piece of knowledge to recreate or a heuristic defining acceptability. 
Thus, by changing values of $F$ and $\mathrm{A}$ an optimization approach can search a space of possible combinations of some given set of features.
Optimizing these variables represents the transformation function in Conceptual Expansion work.

\begin{figure*}[tb]
  \centering
  \includegraphics[width=0.9\linewidth]{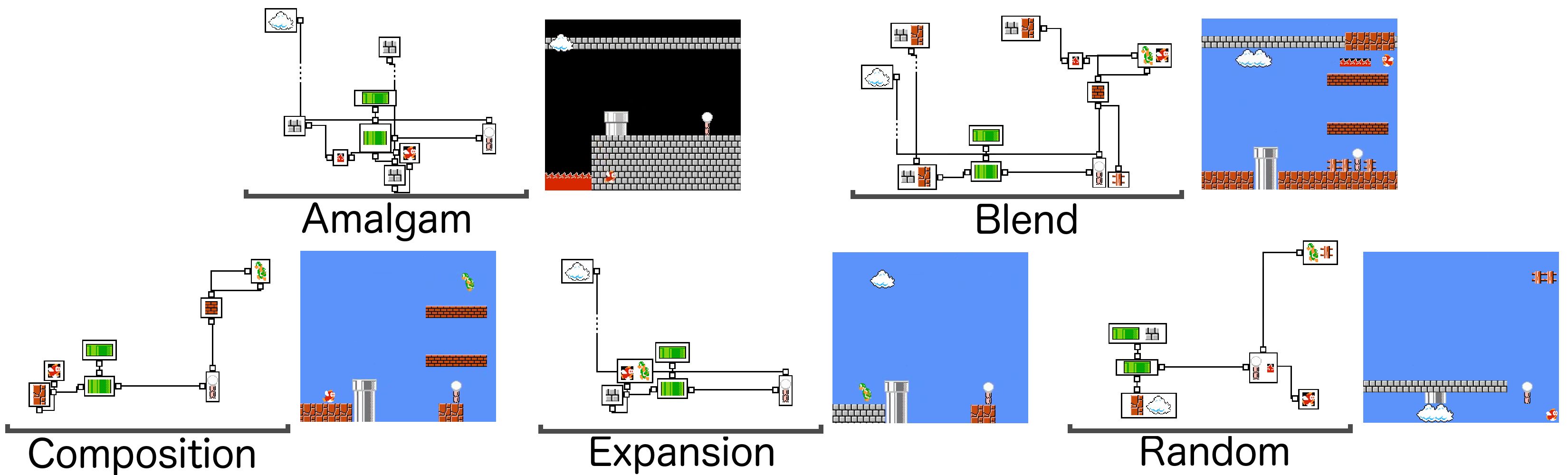}
  \caption{Examples of the different final models and an output level for the different combinational creativity approaches when combining ``castle'' and ``aboveground'' \textit{Super Mario Bros.} levels. Reproduced with permission from \cite{guzdial2018combinatorial}.}
  \label{fig:guzdial2018combinatorial}
\end{figure*}

In \cite{guzdial2018combinatorial}, Guzdial and Riedl applied conceptual expansion to machine learned models of \textit{Super Mario Bros.} level design to produce new types of output levels (e.g., combining boss levels and aboveground levels to create boss-aboveground levels). 
They also made use of conceptual blending and two other combinational creativity approaches as baselines. 
Notably, the levels were not being directly recombined, but instead the input knowledge was learned Bayesian network probabilities that represented level design models. 
After conceptual expansion was applied as the transformation function, the output was a novel Bayesian network that could produce unseen new level types as can be seen in Figure \ref{fig:guzdial2018combinatorial}

In \cite{guzdial2018automated}, Guzdial and Riedl applied Conceptual Expansion to machine learned representations of game rules and game levels for \textit{Super Mario Bros.}, \textit{Kirby's Adventure} \cite{kirbyadventure}, and \textit{Mega Man}. 
This input knowledge took the form of knowledge graphs, combining the game rule and game level knowledge, both derived from gameplay video as the source of the raw knowledge. 
They found that conceptual expansion could outperform conceptual blending and genetic algorithms at recreating a held-out game given the other two games and a partial specification of the held-out game.
However, in this work, Guzdial and Riedl did not attempt to produce fully novel games.
Instead the output knowledge was simply an approximation of a held-out game based on a partial specification of that held-out game.

\begin{figure}[tb]
  \centering
  \includegraphics[width=2.25in]{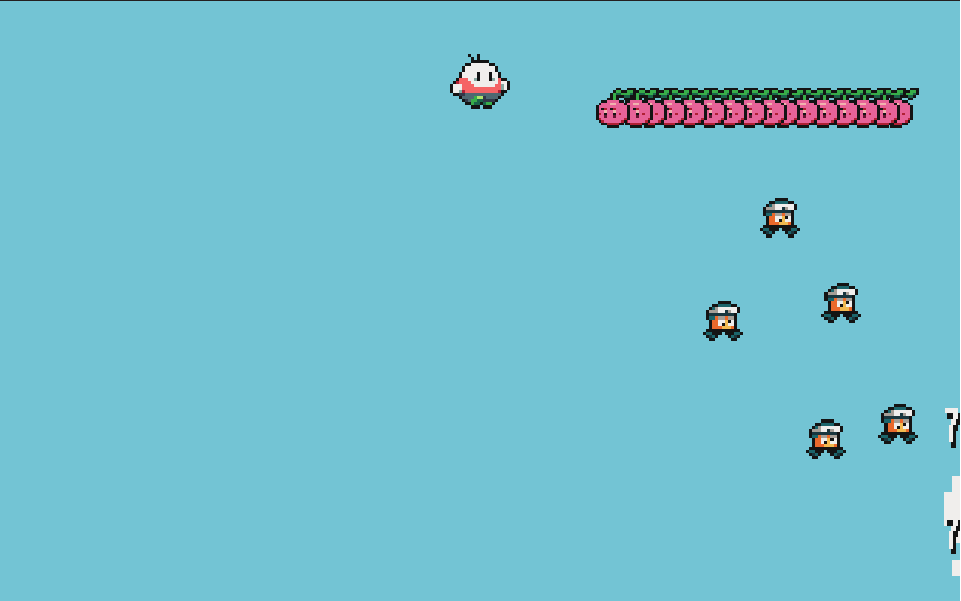}
  \caption{A screenshot of the beginning of one of the Conceptual Expansion games. Reproduced with permission from \cite{guzdial2020conceptual}.}
  \label{fig:guzdial2020conceptual}
\end{figure}

In \cite{guzdial2020conceptual}, Guzdial and Riedl demonstrated the production of novel games from the same input knowledge: the graph of machine learned level and rule information from the three above games. 
They evaluated these games in a human subject study compared to other combinational creativity approaches.
Their transformation function optimized output graphs using a hand-authored heuristic and greedy search over the conceptual expansion representation. 
Their results indicated that the conceptual expansion games outperformed games produced by existing combinational creativity methods, but that the heuristic was poorly designed for game quality. 
Determining how best to guide the search of possible combinations is still an important open problem.
We include a screenshot of one of the games in Figure \ref{fig:guzdial2020conceptual}.

\subsection{Transfer Learning}

Transfer learning is a machine learning technique that focuses on transferring information or knowledge from one domain and applying it in another. In the following sections, we discuss prior PCG-KT methods that have made use of transfer learning as the transformation function.

\begin{figure*}[tb]
  \centering
  \includegraphics[width=0.9\linewidth]{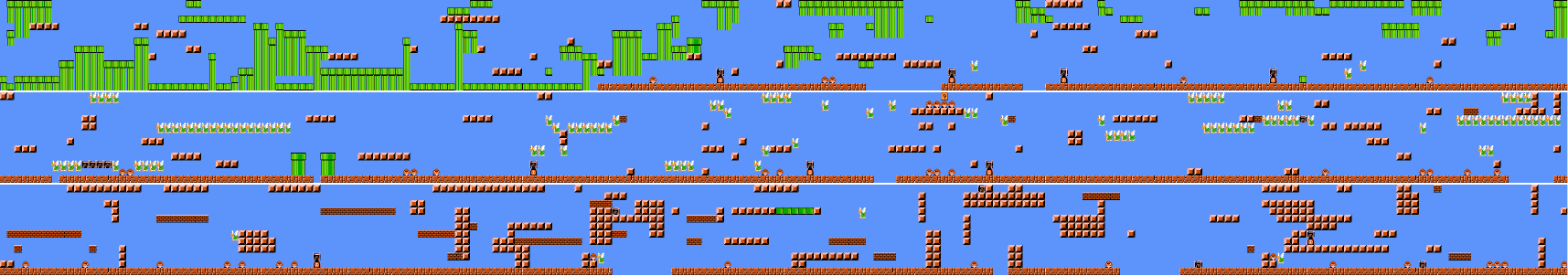}
  \includegraphics[width=0.9\linewidth]{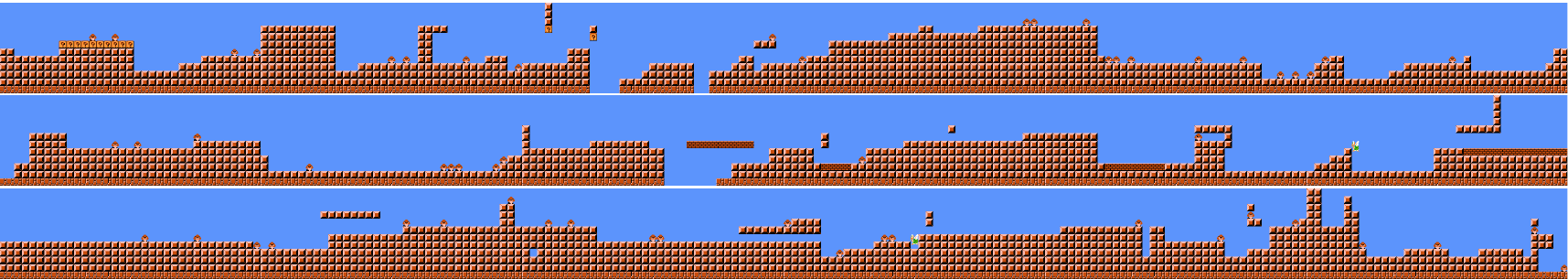}
  \caption{Output levels from domain adaptation Markov models from \textit{Super Mario Bros.} to \textit{Kid Icarus} and \textit{Kid Kool}. Reproduced with permission from \cite{snodgrass2016approach}.}
  \label{fig:snodgrass2016approach}
\end{figure*}

\subsubsection{Domain Adaptation}
This refers to approaches for transforming knowledge extracted from a source domain to supplement or approximate the knowledge of another target domain. The existing domain adaptation approaches for level generation focus on finding a mapping between the representations of the source and target domains.

In \cite{snodgrass2016approach}, Snodgrass and Onta{\~n}\'{o}n presented an approach aimed at automatically defining a mapping between the representations of multiple platforming game level domains.
The core of this approach works by converting the source domain levels to the target domain representation using a subset of mappings between the domains, and filtering out the poorly performing mappings. 
The potential mappings are evaluated comparing frequency of tile occurrences and structures (e.g., $2\times2$ tile blocks) in the transformed input and target domains.
The input knowledge for this approach is the set of tile levels from the input domain.
The transformation function here is the mapping pipeline that tries to define how to map the tiles from the input domain to tiles in the target domain.
After applying this mapping, the output knowledge is a set of levels re-represented as target domain levels.
We include examples of different output levels for different mappings in Figure \ref{fig:snodgrass2016approach}.

\begin{figure}[tb]
  \centering
    \includegraphics[width=\linewidth]{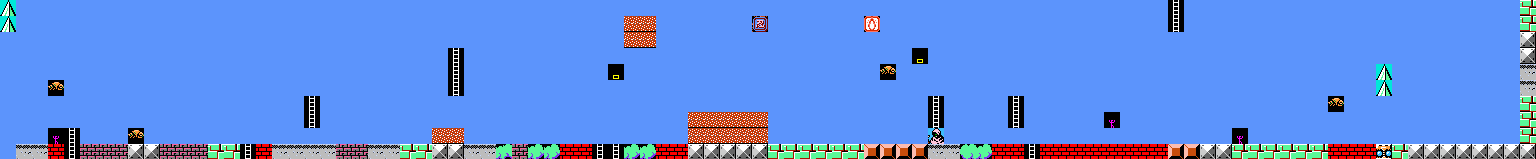}\vspace{.05em}
    \includegraphics[width=\linewidth]{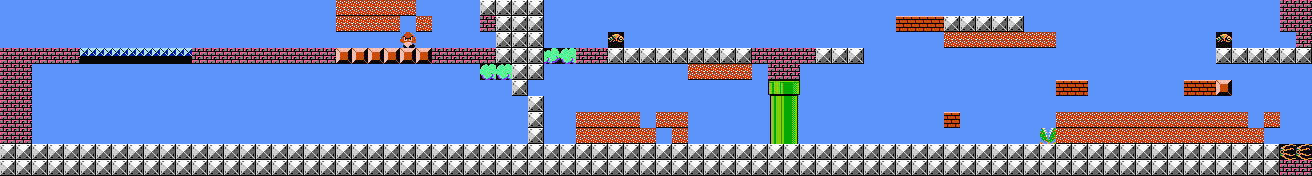}
  \caption{Example \textit{Castlevania} levels produced by the combined domain adaptation and conceptual blending approach. Reproduced with permission from \cite{snodgrass2020multi}.}
  \label{fig:snodgrass2020multi}
\end{figure}

Unlike the above work which aimed to transform from a source domain to supplement a target domain, Snodgrass~\cite{snodgrass2019levels} and Snodgrass and Sarkar~\cite{snodgrass2020multi} aimed to transform a source domain to an intermediate target domain, and then either back to the source domain~\cite{snodgrass2019levels} or to a new secondary target domain~\cite{snodgrass2020multi}. In these approaches, the authors transform the source domain levels into an abstract domain representation meant to capture only the structural properties of the levels from each source domain (i.e., map each tile type to either a solid type or empty type). In \cite{snodgrass2019levels}, they use the abstract levels as input to a level generator that creates variations on that level. The variations are generated by matching the structures in the abstract level with structures in the source domain, and using regions of the source domain levels to fill in the details of the abstract level. In \cite{snodgrass2020multi}, they train VAEs on the abstract levels, and generate new levels in the abstract domain. They then use those generated abstract levels as input to a generator that blends the source domains instead of transforming back to an individual source domain. In this approach the blended levels are created by matching structures in the generated abstract levels with structures across all the source domains, and using the various source domains to fill in the details of the generated abstract level. 
Thus, the input knowledge here consists of the levels in the abstract representation, extracted from the original input levels and the output knowledge consists of the blended levels obtained via applying the transformation function which in this case is the structure matching process described above.
Notice while \cite{snodgrass2020multi} consists of both a domain adaptation stage and a combinational creativity stage, we include it in this section because the former is what enables the latter in this approach.
Figure \ref{fig:snodgrass2020multi} contains example \textit{Castlevania} levels produced with this approach.

\begin{figure}[t]
\centering
\setlength\tabcolsep{1pt}
\begin{tabular}{cccc}
\includegraphics[width=0.12\textwidth]{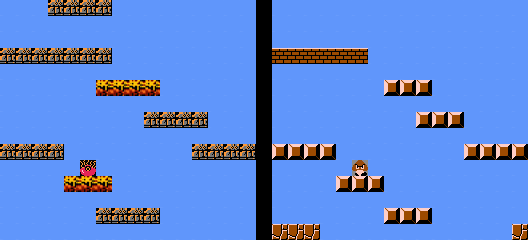} &
\includegraphics[width=0.12\textwidth]{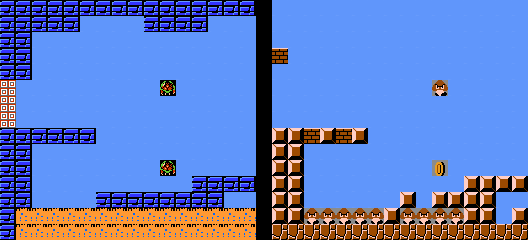} &
\includegraphics[width=0.12\textwidth]{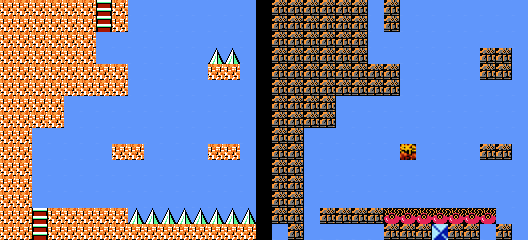} &
\includegraphics[width=0.12\textwidth]{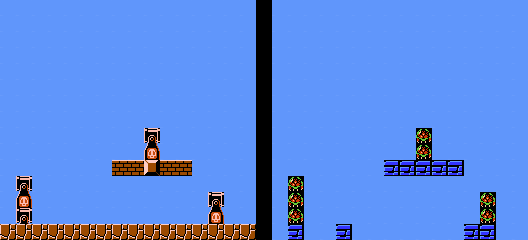} \\
KI-to-SMB & Met-to-SMB & MM-to-KI & SMB-to-Met
\end{tabular}
\caption{\label{fig:tile2tile} Example style-transferred levels involving \textit{Kid Icarus} (KI), \textit{Super Mario Bros.} (SMB), \textit{Mega Man} (MM) and \textit{Metroid} (Met). Reproduced with permission from \cite{sarkar2022tile}}
\end{figure}

\begin{figure}[tb]
  \centering
  \includegraphics[width=0.75\linewidth]{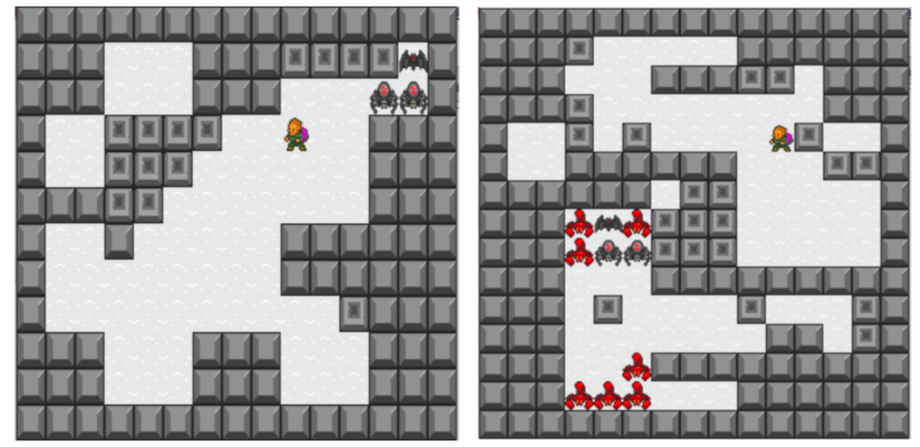}
  \caption{Constructive (left) and search-based (right) generators built from the generalized 3x3s. Reproduced with permission from \cite{beaupre2018design}.}
  \label{fig:beaupre2018design}
\end{figure}

\begin{figure*}[tb]
  \centering
  \includegraphics[width=1\linewidth]{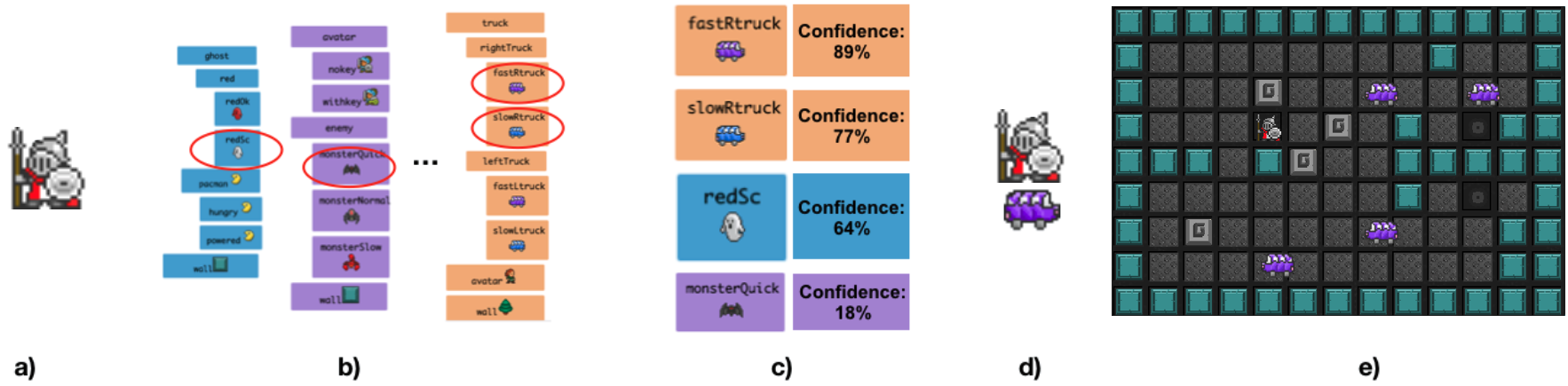}
  \caption{Recommendation process for \textit{Pitako}. Reproduced with permission from \cite{machado2019evaluation}. In this context, all sprites are associated with a game behavior e.g. shooter, chaser, random move, etc. In \textbf{a)}, users provide a sprite to the game engine. In \textbf{b)}, \textit{Pitako} ranks the best sprite matches from other games (red circles) through data mining association (using confidence as a sorting criteria). In \textbf{c)}, users have access to the ranks. In \textbf{d)}, after inspecting the ranks, users select a new sprite. In \textbf{e)}, users can inspect the two sprites in action---the one they provided and the one suggested by the system, imported from another game.}
  \label{fig:machado2019evaluation}
\end{figure*}

\subsubsection{Platformer Style Transfer}
Style transfer \cite{gatys2016neural} is a popular application of creative ML in visual art and has been the subject of a large body of research \cite{jing2019neural}. To implement style transfer for games, Sarkar and Cooper introduced tile2tile \cite{sarkar2022tile}, a method for transferring style between levels of different platformer games. 
In the context of PCG-KT, we can view style transfer in games as extracting knowledge about the style of one game and transforming it into that of another game. While game blending is similar to style transfer, it is distinct. Style transfer translates the style of one domain to that of another where as in blending, the goal is to take the style and/or content of two domains and combine them to create something new. 
The tile2tile approach involves translating levels from a source game into an affordance-based representation using a pre-defined mapping and then using a machine-learned model to convert from the affordance representation to the game-specific tiles of the target game. This approach treats the underlying affordances and the overarching game-specific tile distribution of a level as being respectively analogous to the concepts of content and style in visual art style transfer. 
The input knowledge here consists of the levels of the source game, the output knowledge consists of the same levels style-transferred to the target game and the transformation function is the combination of the hand-authored tile-to-affordance mapping and the model trained to translate from affordances to tiles of the target game. Example style-transferred levels are shown in Figure \ref{fig:tile2tile}.

\subsubsection{Design Pattern Approach for Multi-Game Level Generation}

In \cite{beaupre2018design}, Beaupre et al. took a generalization and selection approach to procedurally generate new levels of particular games by generalizing across level design patterns from multiple games. 
This led to new game levels that did not fit within the set of existing levels, such as Pacman levels with collectibles outside of the default Pacman pellets. 
Their approach took all the game levels from the VGDL and re-represented each level in a more general representation (e.g., all enemies now considered identical, all collectibles now considered identical, etc.). 
From this representation, 3x3 patterns were extracted, which then informed constructive and search-based PCG generators as seen in Figure \ref{fig:beaupre2018design}.
Notably, this approach did not make use of statistical machine learning, but instead utilized an approach similar to case-based reasoning. 
In this work, the input knowledge consisted of the generalized 3x3 design patterns extracted from the original VGDL levels. 
The output knowledge was made up of the generated levels which in turn were built up using the extracted design patterns via an evolution-based transformation function.

\subsubsection{Data mining association for game rule recombination}
In \cite{machado2019pitako}, Machado et al. introduced Pitako, a recommender system for game mechanics for use within their prior game design tool Cicero. The system is built upon the VGDL and GVGAI (General Video Game AI) framework \cite{perez2019general} and uses association rule mining to make suggestions to the designer regarding adding game elements and rules. 
Typically used with transaction databases, an association rule of the form $X \to Y$ says that the presence of an item $X$ in a transaction implies the presence of item $Y$ in the same transaction with some probability. 
To contextualize this in a game scenario, consider \textit{Mega Man X} \cite{megamanx} and \textit{Super Metroid} \cite{supermetroid}. Both have a shooting mechanic and also share the ``hold button for powerful shot" mechanic. That is, if a game implements a shooting mechanic, it may also implement the powerful shot rule. Framed as a transaction, having a shooting mechanic implies the presence of the powerful shooting mechanic in the same game. 
Association rules are discovered using data mining association algorithms such as the Apriori algorithm \cite{agrawal1994fast} to analyze databases of transactions. 
For this purpose, Pitako uses a game catalog. The system uses the games available in the GVGAI framework and applies a depth-first tree search on each of them to read all the game descriptions and put them in the catalog where every game element, from every game, can be disassociated from its original source and easily inserted in a target game. The catalog is organized to facilitate that its items, which are the elements required to describe VGDL games, can be categorized to be used as input for data mining association algorithms. Figure \ref{fig:machado2019evaluation} outlines the Pitako recommendation process. It starts by getting the designer game description set. In (a), the user has only one game element. In (b), the system uses frequent itemset data mining to identify associations that match the element in the designer’s game to elements in the catalog of games. The red ellipses represent elements from different games which are candidates for being recommended. This procedure yields a list of recommendations sorted in descending order by the rule’s confidence level shown in (c). The designer picks the recommendation with the highest confidence level (d) and employs it as a new element in the game. The element shown here is from a Frogger clone and in this example, the user is mixing elements of Frogger and Sokoban. 
When the user picks a recommendation, its game elements are transferred from a game in the catalog and inserted into the in-development game.
Thus, the input knowledge here is the game catalog extracted from the GVGAI games using search while the output knowledge is the designed game which contains elements taken from the input games that the user chose. 
The transformation process involves both the application of the Apriori algorithm on the catalog to generate recommendations as well as the use of these recommendations by users when designing their game. It is interesting to note that the transformation process here is mixed-initiative as opposed to the more automated transformations in other surveyed work.

\section{Discussion}

In this section, we summarize open problems and directions for future work related to PCG-KT. 

\subsection{Evaluation}
It is unclear how to best evaluate knowledge transformation in a generative process. 
Existing works have utilized techniques such as expressive range analysis \cite{smith_analyzing_2010}, playability measures, and user studies, while ML-based systems have used comparisons between training and generated distributions \cite{summerville_expanding_2018}. However, these focus on assessing the goodness and functionality of generated outputs rather than evaluating the properties of the transformation itself. Moreover, in certain cases, even content evaluation methods need to be improved upon. For example, in the case of game and level blending, it remains unclear how best to assess the quality of blended levels and/or levels belonging to an entirely new domain produced by blending. A current common approach is to compare content produced in the new/blended domain with content belonging to the original input domains, however these often signal whether blending or transformation has taken place rather than provide information about the extent, type and quality of blending and transformation. 
This is analogous to our concept of measuring the content distance of the transformation function introduced above, but tells us little about the other properties of the transformation.
Thus in the future, more robust and informative evaluation techniques need to be developed. This would likely require defining specific metrics capable of assessing various properties of knowledge transformation, perhaps utilizing some of the other features highlighted in our framework definition.

\subsection{Extending to Multiple Game Genres}
A majority of works utilizing knowledge transformation (and PCG in general) have focused on platformer games. 
While using games of one genre is more convenient, restricting to one genre limits the scope and possibility space of the transformation process. 
Extending knowledge transformation approaches to multiple genres and domains could enable the generation of new types and varieties of content. Understandably, working with knowledge extracted from multiple genres would pose new challenges in terms of developing techniques for reasoning with such knowledge and applying transformations in a meaningful way. However, being able to do so could promise novel gameplay experiences. Could we design transformation functions able to combine extracted knowledge from a platformer such as Mario with extracted knowledge from an adventure game like Zelda? 
As previously discussed, Sarkar and Cooper \cite{sarkar2021dungeon} took a step in this direction by generating levels that combined platformer segments with dungeon rooms but ignored considerations of mechanics in such levels. Addressing such issues and extracting knowledge from different genres and domains in a manner that is conducive to meaningful combinations and transformations may open up the possibility of building systems that could blend genres of games rather than just levels or games. For example, could such a system combine the mechanics of a platformer with the lock-and-key progression of an adventure game to produce a metroidvania? 
In terms of the PCG-KT framework, such a system would exhibit transformation properties of even higher content distance between input and output knowledge since the output domain of the transformation function would not only be a new game but possibly a new genre altogether.

\subsection{Combining Models and Techniques}
In all previously discussed works, the knowledge transformation process was performed using one primary model or architecture such as Markov chains, Bayes Nets and VAEs. In the future, it would be interesting to consider hybrid knowledge transformation processes that combine the use of these models in some meaningful way. The choice of model used as the derivation function dictates the form of the extracted knowledge. We have seen this take the form of latent vector representations in the case of VAEs and probability tables in the case of Markov chains, when applied to similar raw knowledge. Thus, it might be fruitful to explore how these different extracted knowledge representations could be combined to inform and influence the transformation process and obtain different varieties of output knowledge. For example, could the models used for conceptual expansion be combined with the Markov models used for domain adaptation or with VAEs to learn a continuous latent space of game graphs? Alternatively, could conceptual expansion be applied on latent vectors of different games or the design patterns of some other prior works? Such model ensembles could be viewed as hierarchical knowledge transformation where the knowledge extracted from one derivation function is used as the input to another. In other words, this can be seen as replacing the single derivation function $D$ with a composition of derivation functions $D_n \circ D_m \circ ... \circ D_b \circ D_a$, whose application produces intermediate extracted knowledge $\mathbb{K}_a, \mathbb{K}_b ... \mathbb{K}_m, \mathbb{K}_n$ at each step. Examining the features of such successively extracted knowledge and how they might influence the transformed knowledge could be another fertile direction for future work. Similarly, one could conceive of a composition of transformation functions $T_n \circ T_m \circ ... \circ T_a$ that performs successive transformations on some given extracted knowledge, producing new transformed knowledge at each step. This could take the form of a system that produces game blends (via conceptual blending) and then transfers the style of the blended game to another game (via domain transfer).

\subsection{Design Tools and Controllability}
An important direction for future work is to add more controllability to knowledge transformation processes as well as to democratize PCG-KT models via user-facing designer tools. For example, consider the blending related works discussed earlier. While these enable controllability via specifying objective functions and supervised ML models, they require experience and familiarity with ML and evolutionary algorithms, thereby making it feasible for only researchers and experts to work with such methods. Additionally, several methods discussed in this paper, particularly those based on ML, can be viewed as black boxes that take in input (and optionally a set of constraints) and produce an output without affording much room for user interaction with the generative/transformation process, thereby allowing little authorial control. One could imagine future design tools addressing such concerns by, for example, visualizing the latent design spaces learned by these models and allowing users to interactively navigate such spaces and steer the generation process. Overall, future work needs to focus on building tools that enable a wider variety of users to control such models (e.g., enabling the use of UI elements and interfaces rather than defining one-hot encoded labels and coding up fitness functions for evolution). Such tools could take inspiration from the large body of existing co-creative systems \cite{yannakakis2014mixed}, 
particularly those based on ML, such as Morai Maker \cite{guzdial2018co} and TOAD-GUI \cite{schubert2021toad}, both for making Mario levels, the tool from Schrum et al. \cite{schrum2020interactive} for interactively exploring the latent spaces of GANs trained on Mario and Zelda, as well as Lode Encoder \cite{bhaumik2021lode}, a VAE-based tool for generating \textit{Lode Runner} \cite{loderunner} levels. Note that in all these examples, the tools only work in one game/domain and thus cannot be considered as tools for PCG-KT. In fact, the Cicero and Pitako tools discussed previously are among the few examples of PCG-KT design tools, thus speaking to the need for developing more of these in the future. Sarkar and Cooper \cite{sarkar2020gdcml} provide a blueprint for such a tool but at the time of writing, no implementation exists. Incorporating more controllability into PCG-KT approaches and operationalizing them via user-friendly tools could further advance the field and help PCG-KT methods be applied for game and level design more widely.
In addition to tools that help less experienced users, future works could also consider tools that incorporate user knowledge in performing the transformation process. These could be viewed as creativity support tools \cite{shneiderman2007creativity} where the user enters input knowledge taken from various domains and the tool enables the user to either choose from a variety of automated derivation functions to apply, create their own hand-authored derivation functions or highlight aspects of the input knowledge that are amenable to transformation. Such tools could thus help incorporate a user's creative insights into the PCG-KT process and foster a more co-creative, human-centered form of PCG-KT as opposed to the more automated methods that we have surveyed in the paper.

\section{Conclusion}
In this paper, we introduced Procedural Content Generation via Knowledge Transformation (PCG-KT), a new lens for PCG research that focuses on methods that generate content via transforming knowledge between domains. We offered a formal definition for PCG-KT, outlined a framework for characterizing works under this lens, discussed several examples of existing works that fall under the framework and concluded with directions for future work. Prior survey papers on search-based PCG \cite{togelius2010search} and PCGML \cite{summerville2018procedural} have been successful in highlighting a specific family of PCG approaches as well as pointing out future work directions which inspired further research. For example, several works highlighted in this paper were directly influenced by suggestions made in the PCGML paper. We hope to achieve a similar result with this paper. By highlighting this new branch of PCG, we wish to stimulate further research in this interesting and exciting new direction.


%


\ifCLASSOPTIONcaptionsoff
  \newpage
\fi



\bibliographystyle{IEEEtran}
\bibliography{main}
%

%








\end{document}